\documentclass{article}
\pdfoutput=1

\PassOptionsToPackage{numbers, compress}{natbib}

\usepackage[preprint]{neurips_2022}

\usepackage[utf8]{inputenc} 
\usepackage[T1]{fontenc}    
\usepackage{hyperref}       
\usepackage{url}            
\usepackage{booktabs}       
\usepackage{amsfonts}       
\usepackage{nicefrac}       
\usepackage{microtype}      
\usepackage{xcolor}         
\usepackage{graphicx}       
\usepackage{wrapfig}        
\usepackage{subcaption}     
\usepackage{amsmath}        
\usepackage{makecell}
\usepackage{caption}
\usepackage{placeins}

\title{HashEncoding: Autoencoding with Multiscale Coordinate Hashing}

\author{%
  Lukas Zhornyak$^*$\\
  University of Pennsylvania\\
  \texttt{zhornyak@seas.upenn.edu} \\
  \And
  Zhengjie Xu$^*$ \\
  University of Pennsylvania \\
  \texttt{xzj1999@seas.upenn.edu} \\
  \AND
  Haoran Tang$^*$ \\
  University of Pennsylvania \\
  \texttt{thr99@seas.upenn.edu} \\
  \And
  Jianbo Shi \\
  University of Pennsylvania\\
  \texttt{jshi@seas.upenn.edu} \\
  \phantom{\thanks{~Equal contribution.}}
}

\begin{document}

\maketitle

\vspace{-25pt}
\begin{abstract}\vspace{-5pt}
    We present HashEncoding, a novel autoencoding architecture that leverages a non-parametric multiscale coordinate hash function to facilitate a per-pixel decoder without convolutions. By leveraging the space-folding behaviour of hashing functions, HashEncoding allows for an inherently multiscale embedding space that remains much smaller than the original image. As a result, the decoder requires very few parameters compared with decoders in traditional autoencoders, approaching a non-parametric reconstruction of the original image and allowing for greater generalizability. Finally, by allowing backpropagation directly to the coordinate space, we show that HashEncoding can be exploited for geometric tasks such as optical flow. 
\end{abstract}

\section{Introduction}\vspace{-5pt}
\begin{wrapfigure}{r}{0.55\linewidth}
    \vspace{-6em}
    \captionsetup{justification=centering}
    \begin{subfigure}{0.45\linewidth}
        \includegraphics[width=\linewidth]{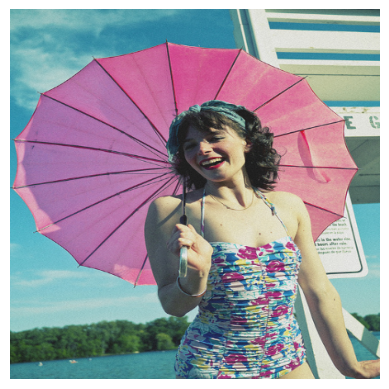}
        \caption{Original \\ {\ }}
    \end{subfigure}%
    \begin{subfigure}{0.45\linewidth}
        \includegraphics[width=\linewidth]{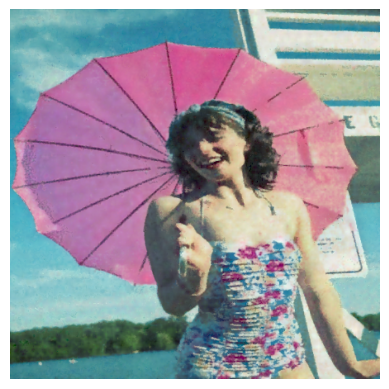}
        \caption{Per-Image Hash Table + Per-Image Decoder}
    \end{subfigure}
    \begin{subfigure}{0.45\linewidth}
        \includegraphics[width=\linewidth]{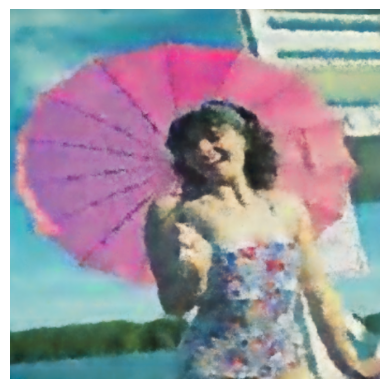}
        \caption{Hash Estimating Encoder \\+ Universal Decoder}
    \end{subfigure}%
    \begin{subfigure}{0.45\linewidth}
        \includegraphics[width=\linewidth]{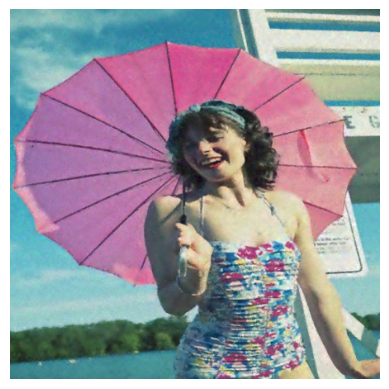}
        \caption{Fine-tuned Encoder Prediction\\+ Universal Decoder}
    \end{subfigure}
    \captionsetup{justification=justified}
    \caption{\textbf{A comparison of the quality} of different optimization strategies with $T = 2^{12}$.}
    \label{fig:quality}
    \vspace{-2em}
\end{wrapfigure}

Instant NGP\cite{muller2022instant} showed near-perfect image and volumetric reconstruction using a highly compressed hash table representation.  All we need to do is:  a) map the pixel spatial coordinate into a hash entry, b) lookup the hash feature values stored, and c) run a simple neural network to `decode' the pixel value.   Diving slightly into the details, we don't directly index each pixel, but their four nearest grid points at multiple image scales.  The hash values at those grid points are bilinearly interpolated before running the decoder.  The magic is that even with a significant hash collision, the simple neural network, along with multi-resolution presentation, can produce near-perfect image reconstruction.

We ask the question: can we view this hash representation and the pixel coordinate-based reconstruction as an auto-encoder mechanism?   To be more precise, given any image $I$, can we compute a perfect hash table $F(I)$, such that when we map a pixel $(x,y)$ to hash entry $u=H(x,y)$, such as $F(I)(u)$ can be decoded via a `universal decoder' $D: D(F(I)(u)) = I(x,y)$?
Note that the hash table $F(I)$ can be thought of as a non-parametric representation of $I$.   

The universal non-parametric can open up new image analysis and synthesis possibilities.  Unlike traditional auto-encoders, our hash-based decoder is minimal, making it significantly easier to conduct backpropagation to the encoding space.  Taking the backpropagation further along the computation path, we can directly optimize spatial functions such as optical flow and 3D structure from motion because the decoder is conditioned on the pixel coordinates.

Learning a hash table and decoder for each image instance is achievable \cite{muller2022instant,lefebvre2006perfect} via direct backpropagation or search.  As we scale up the number of images, we notice the reconstruction via a universal decoder becomes increasingly tricky.  Empirically, for 4,000 Coco images, the reconstruction tends to degrade, first losing ability to reconstruct colors, then the entire image. 

To gain insights into what makes learning the universal hash-based decoder difficult, we analyze the pattern of hash collision.  We find the hash collision is nearly regular: pixels at a fixed spatial interval are mapped into the same hash table entry.   One can visualize such hash function as space-folding: folding an image recursively into a small patch. 
This space-folding property of a hash function shows both the challenges and opportunities of learning a universal decoder. We will show that overcoming these challenges requires changing the way we search for the hash feature values and replacing the bilinear interpolation in the non-parametric sampling step. 

The challenge is that pixels folded onto each other, those mapped into the colliding hash entry, can be very different, making it hard to find a consensus.  However, it is known that natural images have recurrent image statics across multiple image scales, and there are recurring texture patterns within the same image \cite{torralba2003statistics,burton1987color, field1987relations}.   We conjecture that the learning algorithm will likely find pixel consensus in the folded image patch, if a) it looks more broadly at recurrent patterns across the different image scales and b) it searches in a larger spatial neighborhood for recurring texture patterns.  

To implement the idea of looking for recurrent patterns across image scales, we draw inspiration from Transfomer-based image semantic segmentation networks such as SegFormer\cite{xie2021segformer}.  Unlike convolutional FCN based networks, SegFormer uses global attention across space and image scales to detect patterns.  
We use Mix Transformer encoder from SegFormer to find an initial guess of the hash code, significantly simplifying the search for a universal hash-based decoder.  

To implement the idea of searching in a larger spatial neighborhood, we borrow the concept of multiple-sampling used in the work of \cite{jiang2019linearized} that removes sampling bias introduced in the bilinear interpolation.   We also remove a well-known gradient discontinuity inherent in any non-parametric indexing operation by replacing the bilinear interpolation with large multiple-sampling.  This step improves the quality of the learned hash code.  It also allows a more continuous backpropagation signal to the spatial coordinates necessary for optical flow computations. 
 
We demonstrate the following:
1) existence of a universal coordinate based hash auto-encoder,
2) translation-invariance at multiple-level of hash encoding,
3) ability to compute the optical flow using the universal coordinate based hash auto-encoder.

\section{Related Works}\vspace{-5pt}

 We are inspired by the work of \cite{muller2022instant} that demonstrated per-image encoding using a small neural network decoder with a multiresolution hash table of trainable feature vectors.   Stochastic gradient descent is shown to be effective in such a per-image setting.  However, an extension to a universal decoder and trainable hash features for a potentially infinite number of images is not achievable with this design.  Per-image/volume instance hash optimization has been studied in  \cite{teschner2003optimized, lefebvre2006perfect} where the focus is searching for a perfect hash mapping that minimizes the hash collision.  Our work does not focus on hash table indexing. Instead, we use a fixed hash function that results in a space-folding property. 

We also draw intuitions from natural image statistics literature, \cite{torralba2003statistics,burton1987color,david2004natural, field1987relations, redies2008fractal}.  This body of work showed that 
the characteristic feature of natural images is their scale invariance: when zooming in and out at different image scales, the statistical properties of the Fourier spectral components remain relatively constant.  This fractal-like property is reflected in the empirical finding that the spectral power of natural scenes falls with spatial frequencies. 
We conjunct that recurrent feature statics across multi-scale can be discovered by SegFormer type of network which global attention across space and scales.  We share a similar insight with the works of \cite{zontak2011internal, huang2015single, shocher2018zero} which have demonstrated the power of using multi-scale recurrent image statistics for image processing.

Pixel coordinate based neural network was used in CoordConv \cite{liu2018intriguing} for feature extraction and in COCO-GAN \cite{lin2019coco} for image synthesis.   The work of \cite{stanley2007compositional} showed a neural network could create complex geometrical patterns using purely spatial coordinates.  Two recent works \cite{anokhin2021image,skorokhodov2021adversarial} demonstrated using implicit neural representations, independent per-pixel image synthesis can be achieved without using spatial convolutions.  Our decoder design shared the same property, with a much-simplified image generator network design.

\section{Methods}\vspace{-5pt}

\begin{figure}
    \centering
    \includegraphics[width=\linewidth]{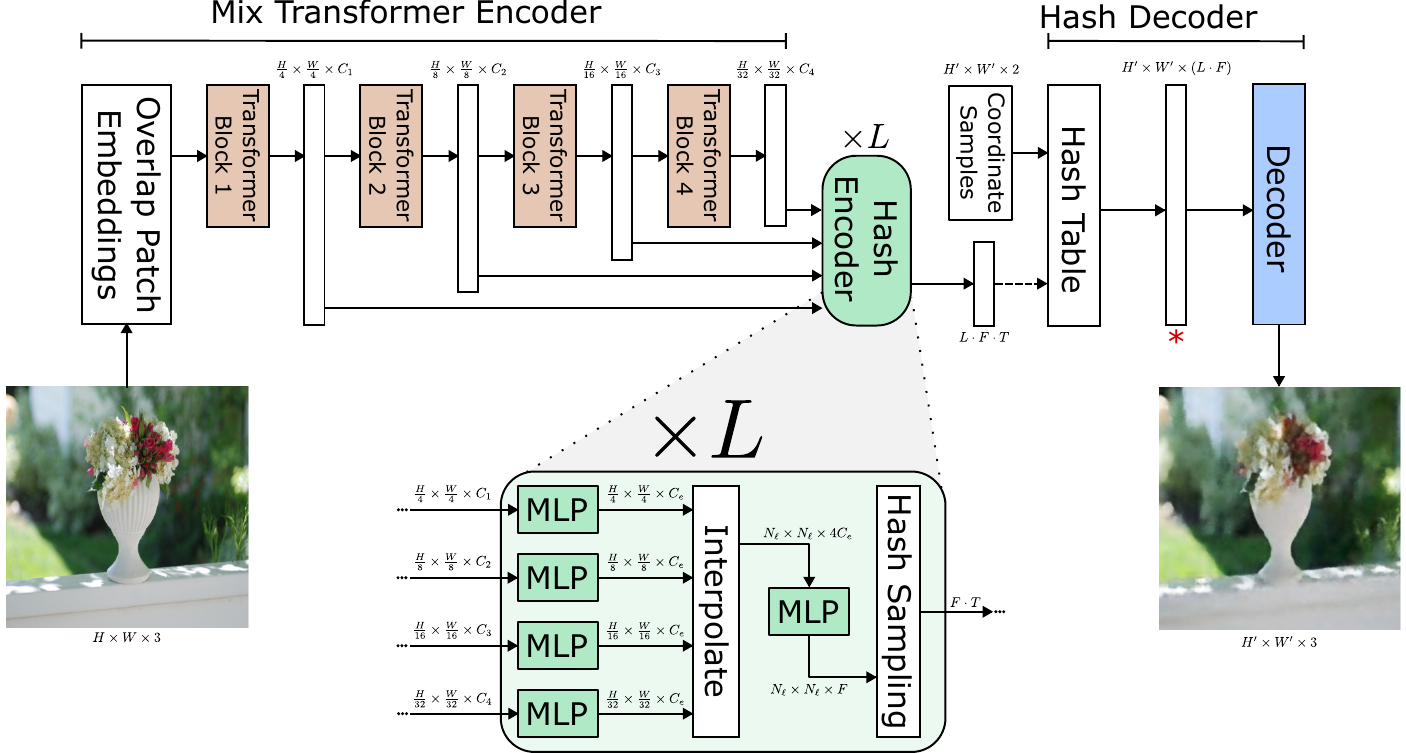}
    \caption{\textbf{The architecture of HashEncoding} contains three components: a Mix Transformer encoder \cite{xie2021segformer} which generates image features at four scales, the proposed hash encoder which interpolates and samples these features to produce $L$ vectors of size $F \cdot T$ that define the hash table, and a hash decoder \cite{muller2022instant} which polls the generated hash tables and decodes the results at the provided coordinates. Here, MLP indicates that operations are performed only across the channels, not spatial coordinates. In the above figure, coloured entries indicate components with learnable parameters.} 
    \label{fig:architecture}
\end{figure}

As shown in Figure \ref{fig:architecture}, the HashEncoding architecture is comprised of three components: a Mix Transformer encoder, a hash encoder, and a hash decoder. A detailed explanation of these components follows.

\subsection{Mix Transformer Encoder}
We first tried directly applying the approach of \cite{muller2022instant} to a large batch of images to find a universal neural network decoder with a multiresolution hash table of trainable feature vectors.   Empirically the Stochastic gradient descent (SGD) often gets stuck in the local minimum resulting in poor image reconstruction: which is not surprising since the hash table structure itself does not provide enough learning mechanism to generalize across images.    We need a network to proactively find patterns across all images at multiple image scales to guide the universal decoder's learning.  

We can think of our design as a particular type of autoencoder, where the decoder is an image coordinate hash-table lookup followed by a tiny pixel decoder.  The purpose of the encoder is to get close to the correct trainable feature vectors so that SGD can `fine-tune' the per-image hash values.  Our autoencoder departs from the traditional design in two aspects.  First, in most autoencoders, dimensions of the embedding space carry no explicit spatial information; each feature dimension represents some component of the semantic information of the \textit{entire} image.   With a multi-resolution hash table (section \ref{sec:hash_table}) however, elements are highly specific to certain regions (or patterns) on the image.  Thus, our encoder must be able to retrieve high-level (coarse) features while preserving low-level (fine) features for all resolutions in the hash table.   Second, while the encoder in most autoencoders attempts to create a bottleneck in the information space to extract useful features, we rely on the spatial compression inherent in the hashing function to produce a bottleneck (section \ref{sec:hash_encoder}).



Our intuition is that the image hash encoder task for learning a hash feature shares a similar challenge to the semantic segmentation task: to find multi-scale recurrent image patterns while preserving fine details from an image.  We first use a 
semantic segmentation network to map the image to its trainable feature vectors without considering the hashing step.   We then compress the feature vectors into the hash table via the known space-folding hashing function.   

Not all semantic networks are equal for our encoding purpose.  
While the convolutional FCN approach is an attractive choice, we find that it failed to discover hash feature patterns, as shown in our experiments (see supplementary materials). 
Our task requires a robust mechanism to extract patterns from spatially distant pixels;
convolution is inherently an operation which only pools local information.
As a result, FCNs struggle in this regard.
As such, we use Mix Transformer \cite{xie2021segformer}, a recent transformer-based architecture, to extract image features.
Throughout the remainder of the paper, we use the MiT-B0 configuration with $(C_1, C_2, C_3, C_4) = (32, 64, 128, 256)$.

An additional advantage of Mix Transformer is that it is inherently multi-scale, providing four levels of feature maps in the selected configuration.
In the following section, we describe how we merge these feature maps to encode the $L$ levels of the multiresolution hash table.

\subsection{Hash Encoder}\label{sec:hash_encoder}
To generate the entries for each level $\ell$ of the hash table, we utilise a structure inspired by the decoder from SegFormer\cite{xie2021segformer}.
Given the four feature maps provided by the Mix Transformer Encoder, we extract $C_e = 16$ features from each using MLPs before interpolating them to a common $N_\ell \times N_\ell$ resolution and concatenating them, where $N_\ell$ is the resolution of the $\ell$-th level of the hash table.
This combined feature map is then passed through an additional MLP to extract $F = 2$ features.

To generate a feature vector for the hash table from this feature map, we need to apply the hashing function on the coordinates of the map to associate each point with an entry in the table.
By generating the feature maps at the resolution of each level of the hash table, we avoid the need to calculate an inverse of the multi-sampling procedure.
Instead, each dimension of the feature vector can be associated to a binary mask on the feature map (see Figure \ref{fig:hash_table}).
In this work, we define the value of each dimension as the mean of all elements selected by this mask.

\subsection{Hash Decoder}
\subsubsection{Hash Table}\label{sec:hash_table}
\begin{figure}
    \centering
    \includegraphics[width=\linewidth]{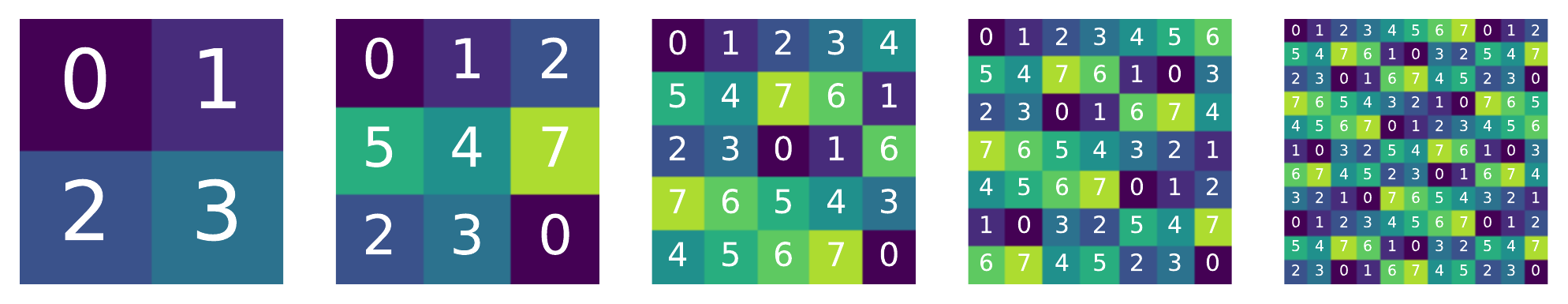}
    \caption{\textbf{Index mapping in the hash table}. We investigate the regularity of the hash table by visualizing the indices mapping to the entries. From coarse to fine and by taking the second hash table from left, we generally observe that the hash table expands to finer resolutions on the basis of previous mapping, and repeat such mapping to form a pattern.}
    \label{fig:hash_table}
\end{figure}

Given an input coordinate $\mathbf{x} \in \mathbb{R}^{d}$ from the image $I$, we are interested in an encoding which exploits the coordinate points and improves the approximation quality efficiently. We approach this goal by constructing a multi-resolution hash table with $L$ independent hash tables of size $T$ as layers, scaling from base resolution $N_{\min}$ to the finest resolution $N_{\max}$ and covering learnable semantics at all levels. Such non-parametric approach achieves better efficiency than neural networks in a way that 1) the $\mathcal{O}(T)$ table lookup at each layer can take place in parallel, and 2) it scales to $N_{\max}$ effectively with $\mathcal{O}(\log(N_{\max}/N_{\min}))$ layers \cite{muller2022instant}. A shallow decoder $\mathcal{D}$ is therefore sufficient enough to reconstruct the image with quality using the concatenation of output features from each layer, which in our case is a two-layer MLP.

At each layer $\ell$, the coordinate $\mathbf{x}$ is scaled by the corresponding resolution $N_{\ell}$, where
\begin{equation}
    N_{\ell}:=N_{\min}\exp\Big((\ln N_{\max }-\ln N_{\min })/(L-1)\Big)
\end{equation}
$\lceil\mathbf{x}\cdot N_{\ell}\rceil$ and $\lfloor\mathbf{x}\cdot N_{\ell}\rfloor$ then span a voxel with $2^d$ vertices which will be mapped to fixed entries of the hash table at layer $\ell$ using a spatial hash function \cite{teschner2003optimized}, and coordinates within the voxel are mapped to the same entries. A spatial hash function can also raise collisions, where different voxel vertices point to the same entry, bringing disparate coordinates together. The hash table implicitly optimizes the collisions since the gradients to each entry will be averaged, and the more important samples will dominate the collision average so that the table entry can reflect its need of the higher-weighted point \cite{muller2022instant}. Such space-folding property of hash table refines and accelerates generalization through collisions, as low-weighted coordinates in the entry learns to prioritize other sparse entries, and high-weighted ones are folded and grouped. In our implementation, we used $N_{min} = 4$ and $N_{max} = 346$, with $L=12$.

Multi-resolution hash table also demonstrates a regularity which implicitly contributes to retaining the scale-invariant statistical properties of a natural image across multiple resolutions as well as the recurring patterns and textures within the image. Figure \ref{fig:hash_table} shows an increasing regularity with the expansions of finer resolutions in the hash table. Starting from the second resolution, the hash table expands along both axes with the base unchanged, and the same local pattern repeats in finer resolutions. 

\subsubsection{Multiple-Sampling}
\begin{figure}
    \centering
    \begin{subfigure}{0.33\linewidth}
        \includegraphics[width=\linewidth]{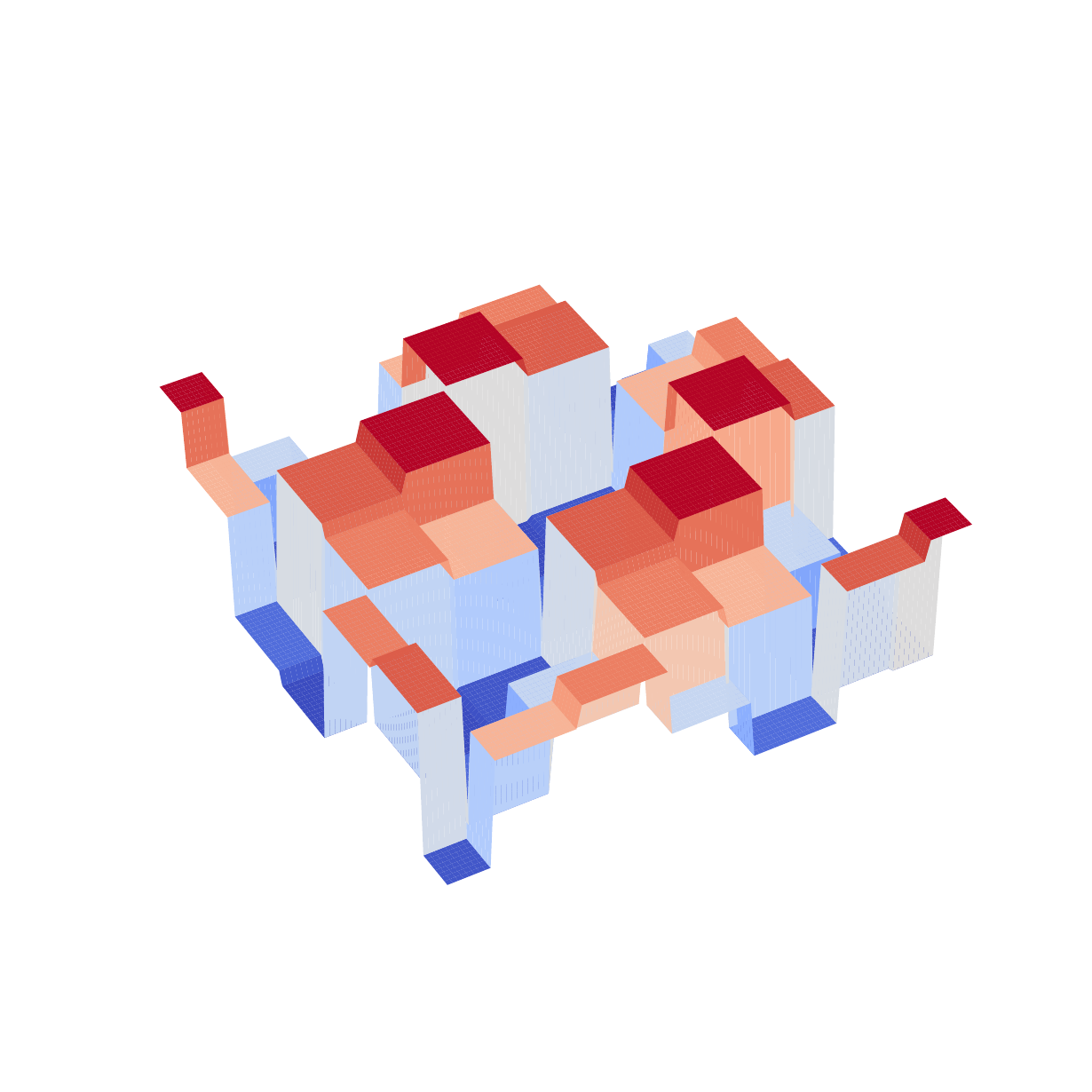}
        \caption{Nearest}
    \end{subfigure}%
    \begin{subfigure}{0.33\linewidth}
        \includegraphics[width=\linewidth]{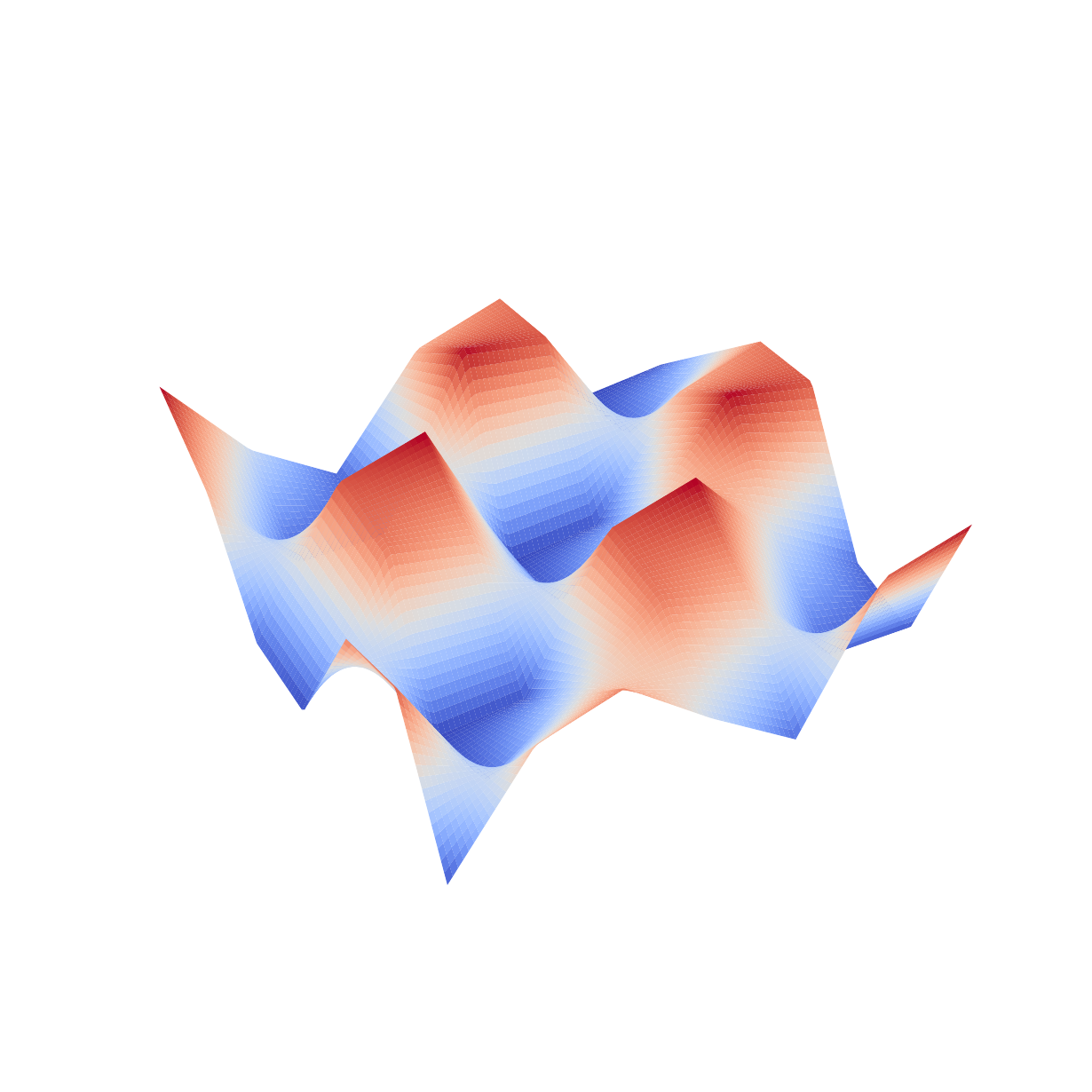}
        \caption{Bilinear}
    \end{subfigure}%
    \begin{subfigure}{0.33\linewidth}
        \includegraphics[width=\linewidth]{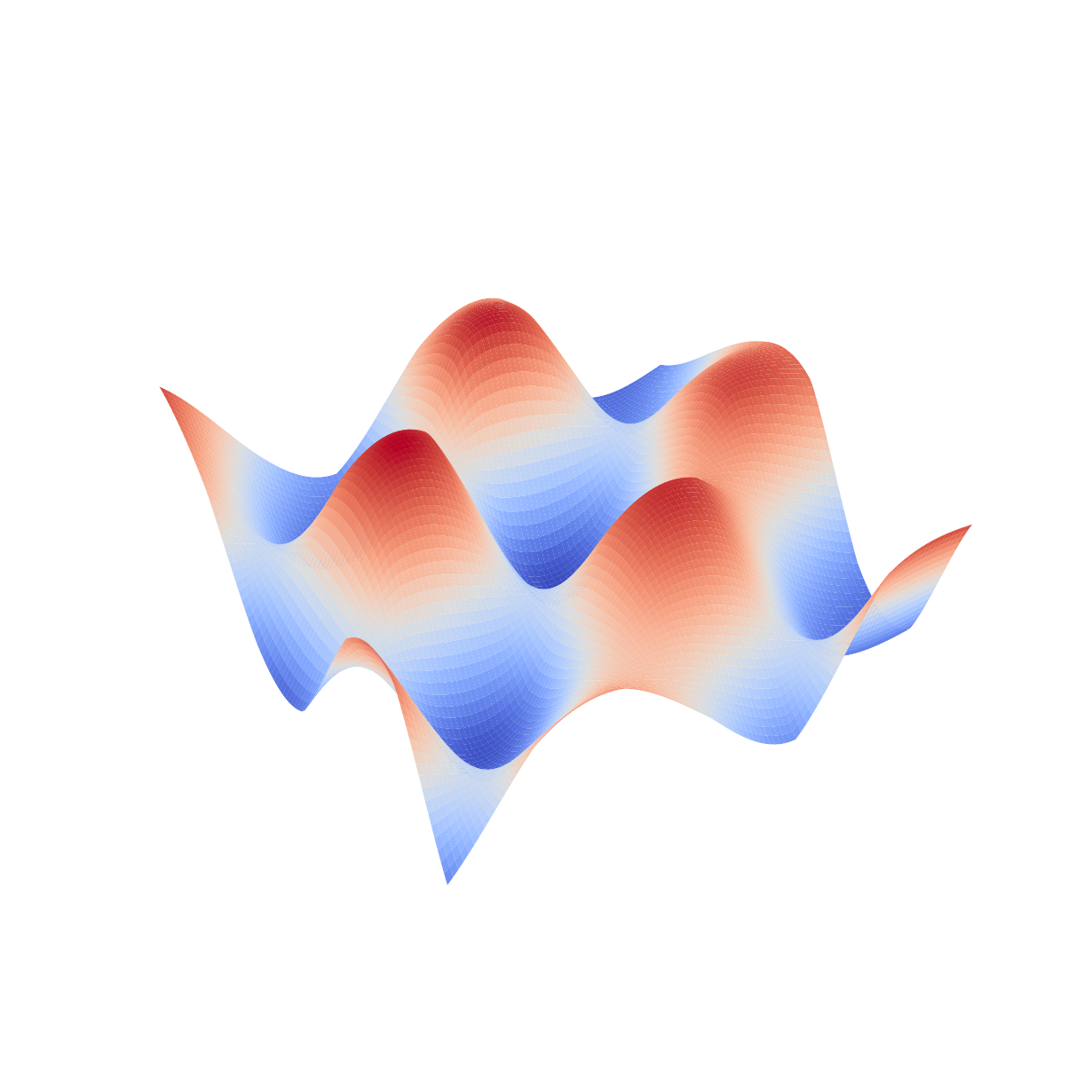}
        \caption{Lagrange Polynomial with k=2}
    \end{subfigure}
    \caption{\textbf{Comparison between different interpolation methods}. With higher orders, Lagrange Polynomial can achieve much smoother interpolation results compared with nearest and bilinear interpolation, even when the data is noisy. Especially at the turns of the data, the sharp discontinuity from both bilinear and nearest can do harm to the gradients and back-propagation, while Lagrange Polynomial can provide better stability and performance.}
    \label{fig:interpolation}
\end{figure}


Extracting the feature of a point in a specific resolution requires interpolation from the pixels nearby. Nearest Neighbor or Bilinear interpolation is a natural choice, and for learning a per-image decoder, this works well.   However, as we will show, the selection of samples used for interpolation would significantly impact the ability to learn a universal decoder and backpropagate to the image coordinates.  Our design calls for sampling multiple sample grid points near a specified image coordinate instead of the 4 nearest grids.   There are two reasons.   First, since many pixels collide at higher-level (fine-resolution) hash tables, the noises for each entry at these levels are high. The nearest neighbor and bilinear grid interpolation, which are not $L_1$ smooth, can potentially amplify the noise in the hash features, as shown in our experiments \ref{fig:invariance}. Second, the backpropagation gradient of the Nearest Neighbor or Bilinear interpolation is highly discontinuous at the sample points.  As ~\cite{jiang2019linearized} points out, back-propagating with respect to coordinates through such small grids can be harmful, especially in the finer-resolutions.  Following the idea of multi-sampling in  ~\cite{jiang2019linearized},  we use additional grid point sampling from coarser-resolution and use bi-Lagrange polynomial with order $2k-1$ to interpolate the pixel feature value.


For illustration purpose, we first show how Lagrange polynomial works for 1-D case. Given $2k$ points with their coordinates in 1-D represented as $x_1, ..., x_{2k}$ and their corresponding values $v_1, ..., v_{2k}$.  WLOG, we assume $x_1 < ... < x_{2k}$, where strict inequality is necessary. Given a coordinate $x$ with $x_1 \leq x \leq x_{2k}$, the interpolated value $v$ at $x$ is given by the following linear combination,
\begin{equation}
    v = \sum_{i=1}^{2k} v_i L_i(x), L_i(x) = \prod_{j\neq i}\frac{x - x_j}{x_j - x_i}
\end{equation}
Extention of 1-D interpolation to 2-D is similar to how bilinear interpolation improves upon linear interpolation: using values at $x$ for $y_1, ..., y_{2k}$, we can use second Lagrange polynomial interpolation along $y$-axis. In our setting, we specifically have $x_k \leq x \leq x_{k+1}$ as we want the same number of pixels smaller and greater than the grid coordinates. By doing so, we achieve smoother transitions between intervals compared to bilinear interpolation (see Figure \ref{fig:interpolation}). This brings smoother gradients across edges or noisy regions. Note that when $k=1$, Lagrange polynomial is just a bilinear interpolation.

Examing the trade-offs of Lagrange polynomials with different $k$ values, as $k$ increases, the total number of pixels used by interpolation increases by $k^2$.   A larger $k$ provides the image pixel with more global information to optimize and flatten the noises within the hash tables, while using more time and memory.  We find $k=2$, which uses 16 points to interpolate, is a good trade-off. 
\section{Experiments}\vspace{-5pt}
\subsection{Reconstruction Quality}\label{recon_quality}
We hypothesize that there exists a small universal decoder which, given the appropriate hash table, can regenerate an image using only coordinate information.
This is shown in Figure \ref{fig:quality}.
While the encoder's initial estimate of the feature vector that defines the hash table shows significant noise, 100 steps of fine tuning results in a reconstruction that approaches the optimal reconstruction given the chosen structure of the hash table.

\begin{figure}
    \centering
    \begin{subfigure}{0.3\linewidth}
        \includegraphics[width=\linewidth]{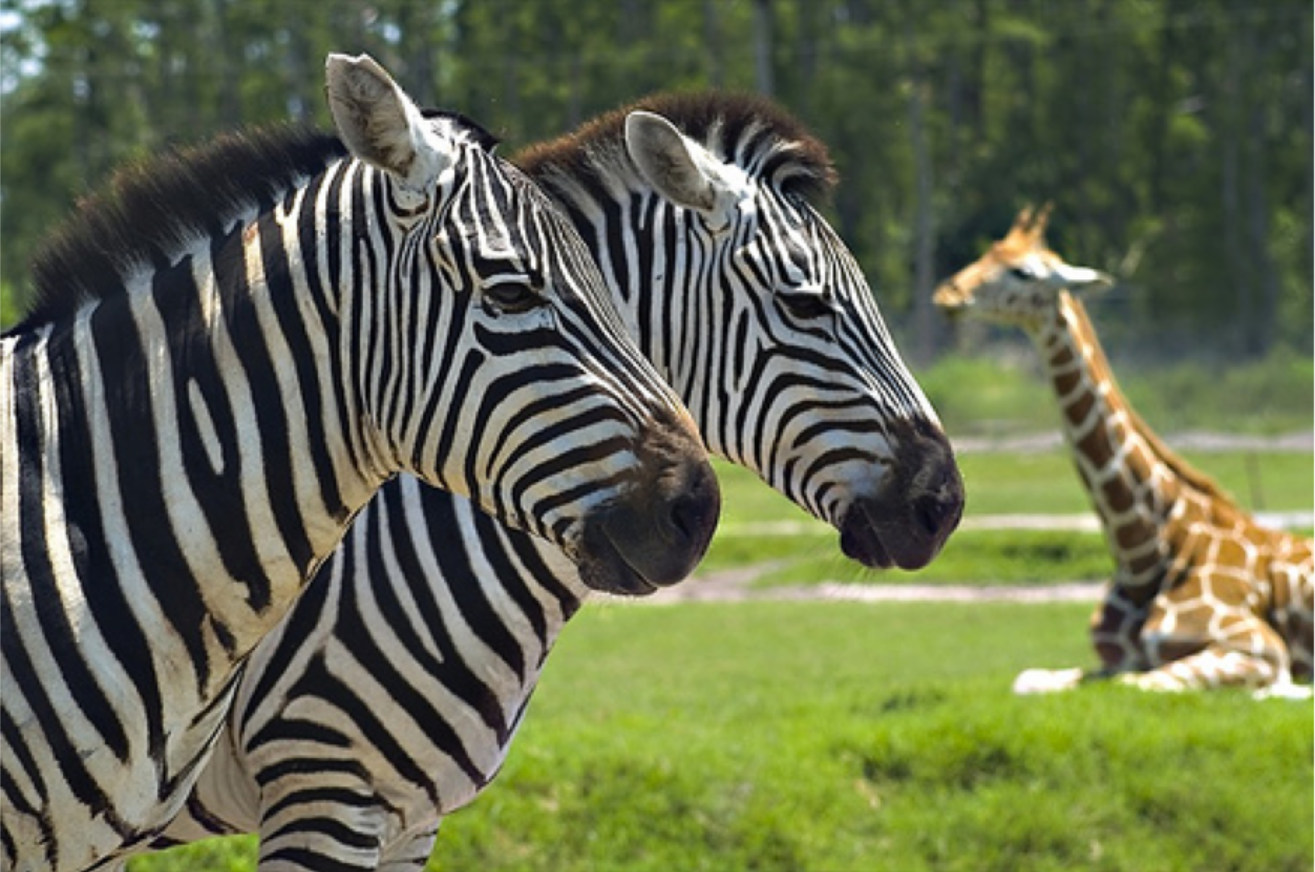}
    \end{subfigure}%
    \begin{subfigure}{0.3\linewidth}
        \includegraphics[width=\linewidth]{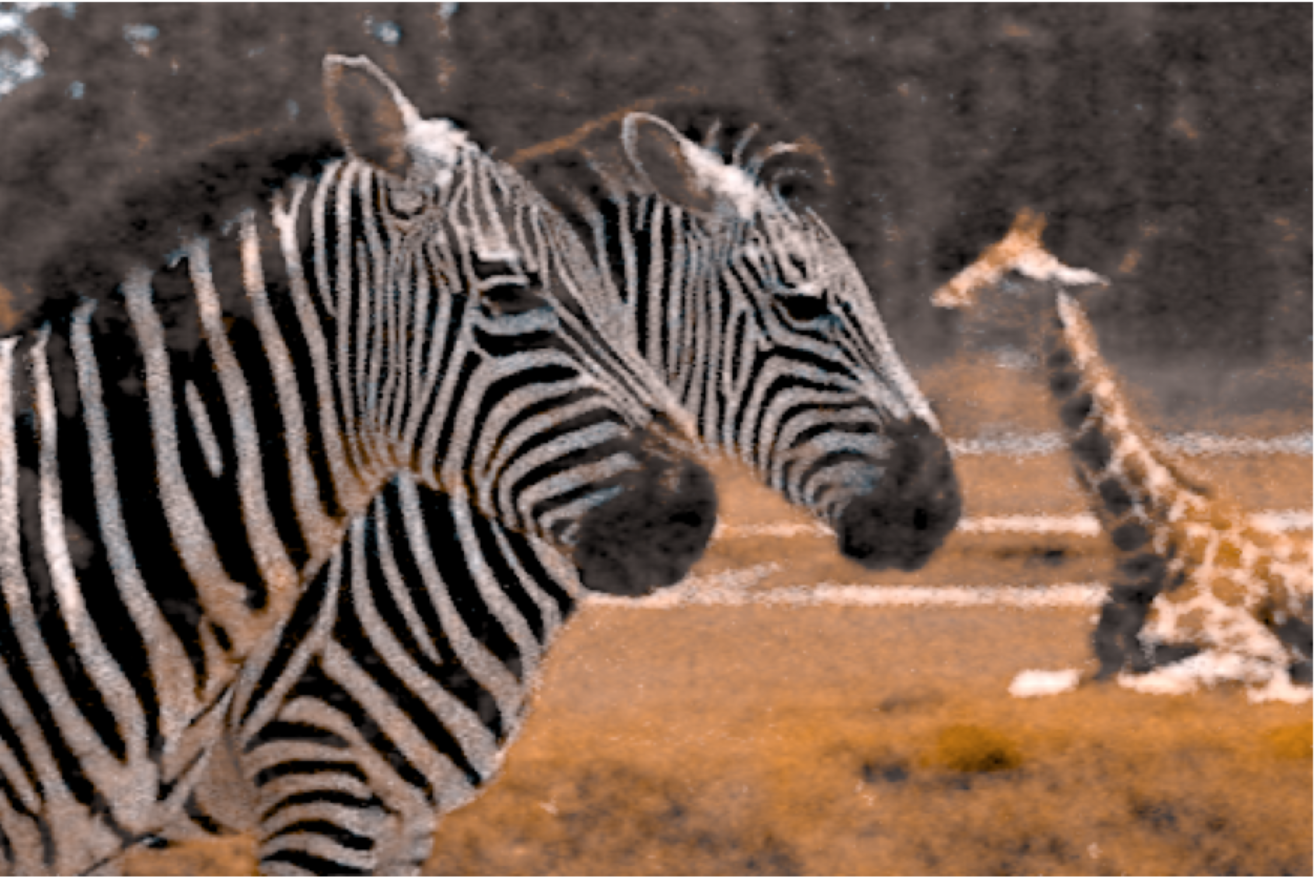}
    \end{subfigure}%
    \begin{subfigure}{0.3\linewidth}
        \includegraphics[width=\linewidth]{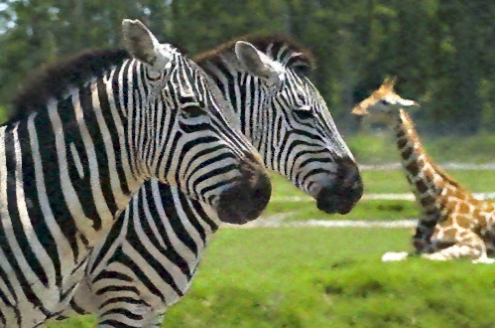}
    \end{subfigure}
    \begin{subfigure}{0.3\linewidth}
        \includegraphics[width=\linewidth]{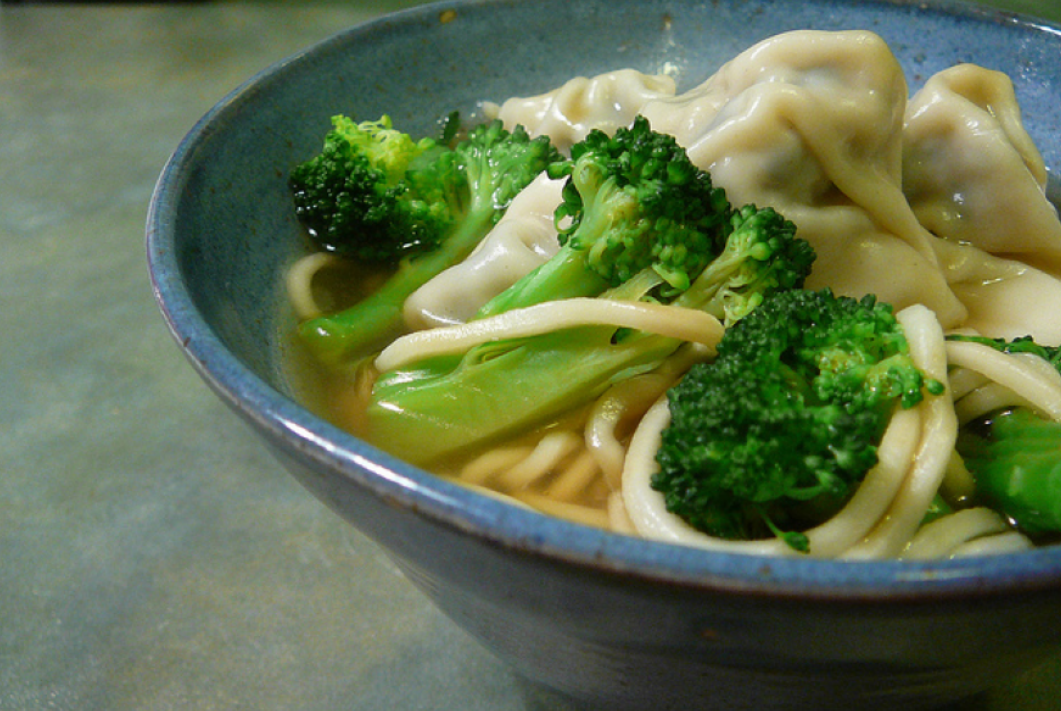}
    \end{subfigure}%
    \begin{subfigure}{0.3\linewidth}
        \includegraphics[width=\linewidth]{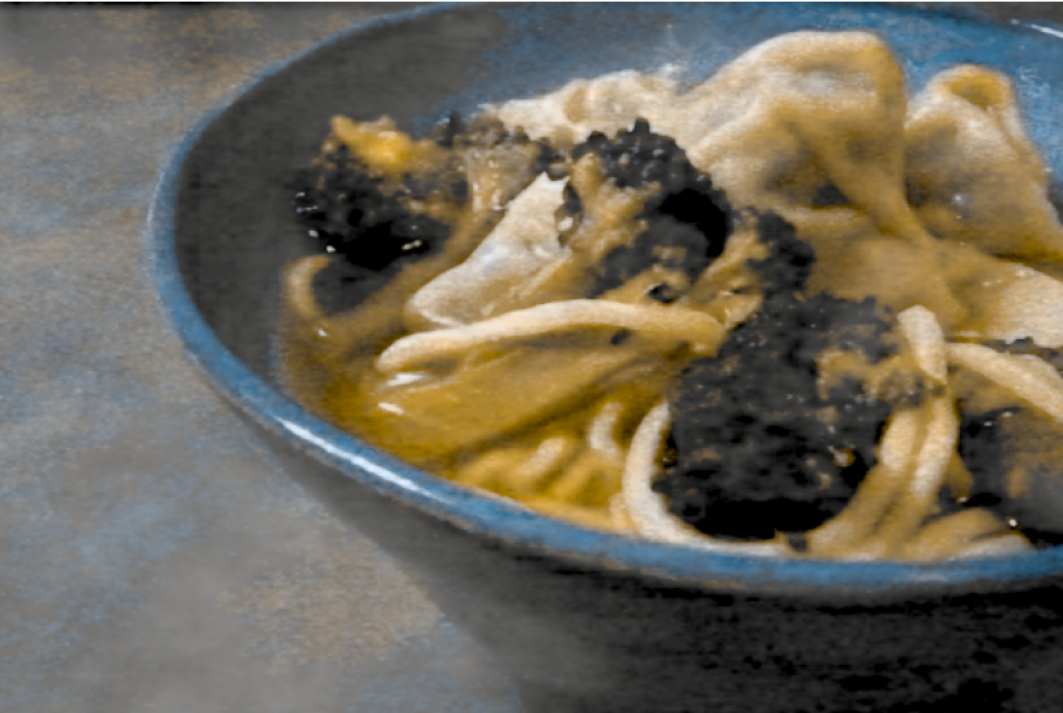}
    \end{subfigure}%
    \begin{subfigure}{0.3\linewidth}
        \includegraphics[width=\linewidth]{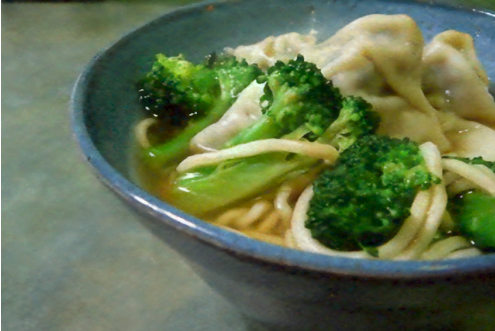}
    \end{subfigure}
    \begin{subfigure}{0.3\linewidth}
        \includegraphics[width=\linewidth]{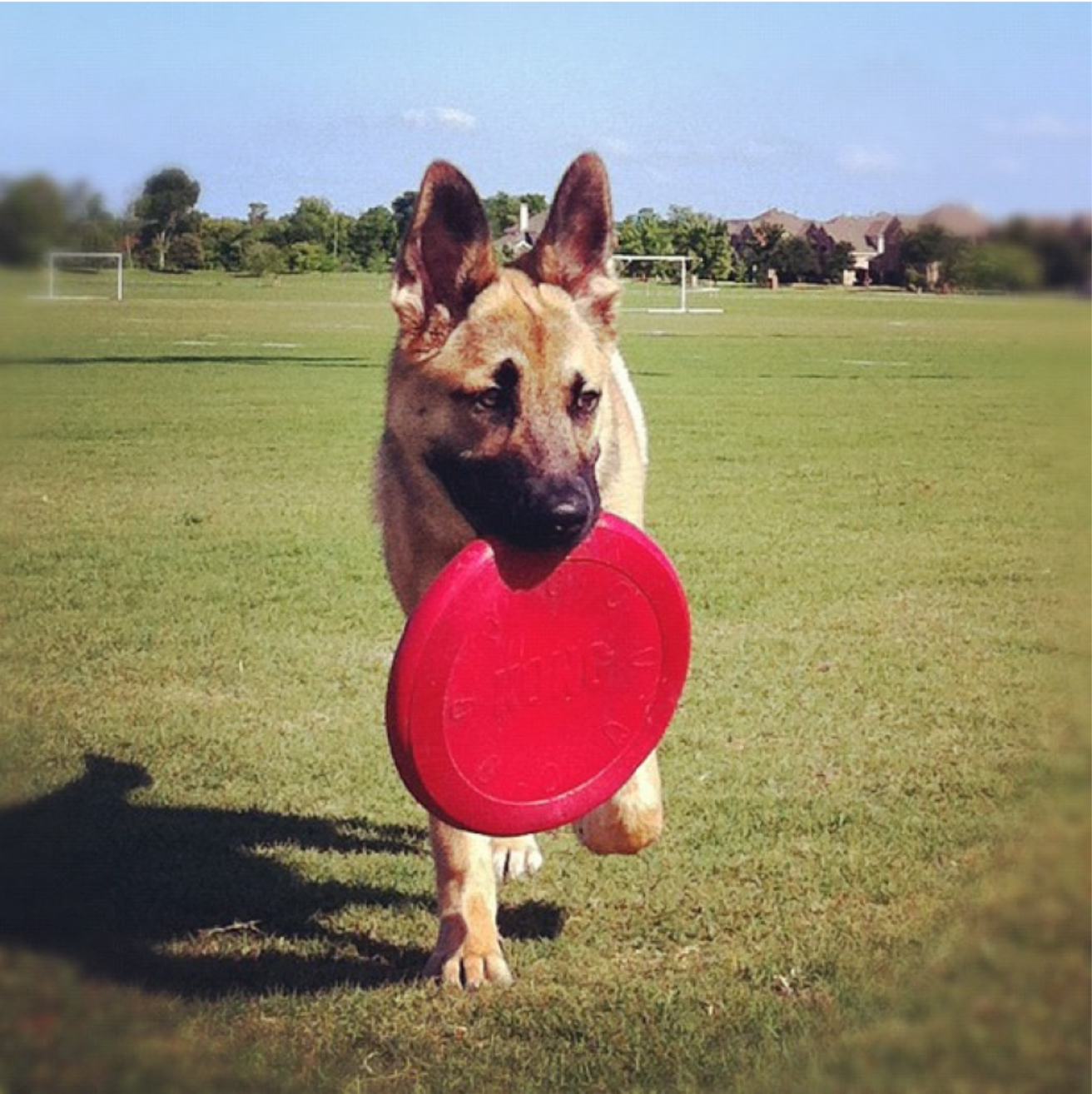}
    \end{subfigure}%
    \begin{subfigure}{0.3\linewidth}
        \includegraphics[width=\linewidth]{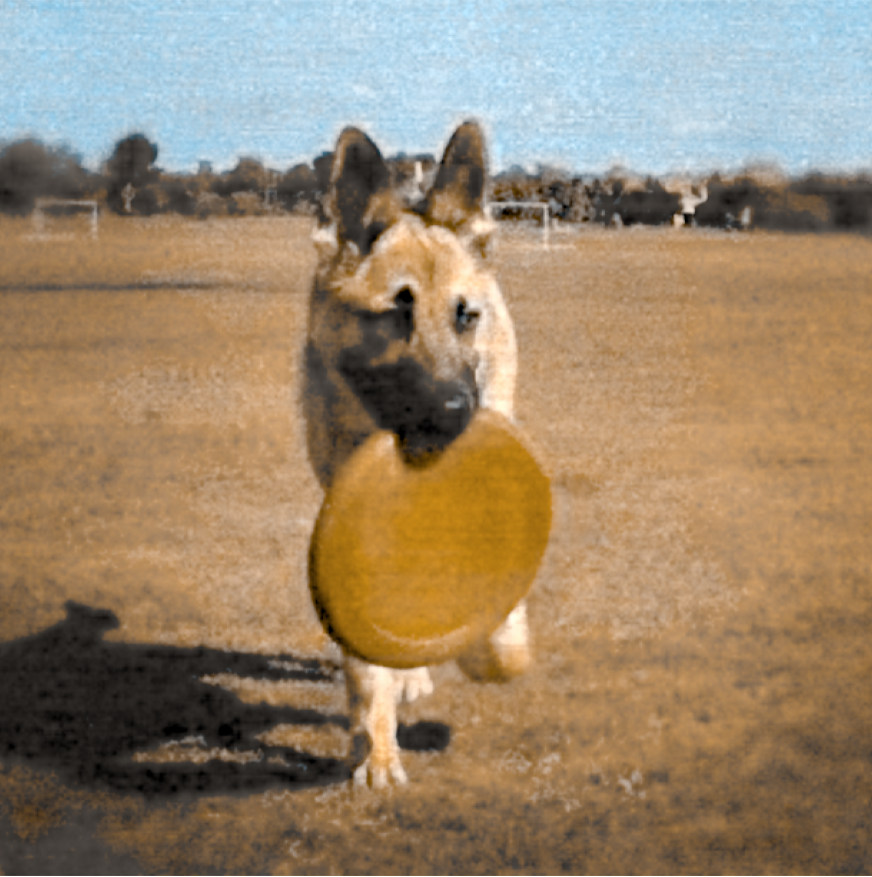}
    \end{subfigure}%
    \begin{subfigure}{0.3\linewidth}
        \includegraphics[width=\linewidth]{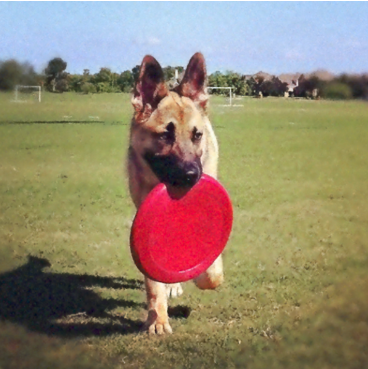}
    \end{subfigure}
    \caption{\textbf{Dimensional loss} observed in direct optimization of hash tables for a batch of images with a shared decoder. The left column is the original image, the middle is the reconstructed image from optimizing the hash table and a shared decoder for a batch of images, and the right column is the reconstruction produced by fined-tuning the prediction produced by the hash estimating encoder.}
    \label{fig:degrade}
\end{figure}

One might question what the purpose of the encoder is if additional fine-tuning is necessary.
If the goal is to produce a universal decoder, why not instead optimize the feature vector defining the hash table directly?
Figure \ref{fig:degrade} shows that this is not an effective strategy.
As the number images in the batch increases, the shared decoder first loses the ability to accurately reconstruct some colours, then all colours, then the image entirely.
We hypothesize that hash estimating encoder ensures that the feature space the decoder operates is constant across images.
Thus, less of the capacity of the decoder is tuned for specific images and more is available for accurate reconstruction.

To quantitatively demonstrates the effectiveness of the hash estimating encoder, coupled with short fine tuning and the improvements provided by multisampling, we calculate the LPIPS score \cite{zhang2018unreasonable} of various optimization strategies against the original image.
By using deep image features provided by AlexNet, the LPIPS score provides a widely used measure of the perceptual quality of a given reconstruction.
The scores are summarised in Table \ref{tab:percep_loss}.

For direct optimization, we train a unique decoder and hash table for each image for 1000 steps; this represents an upper bound on the reconstruction ability given the structure of the hash table and decoder. 
This is compared against the results from decoder trained on all image for both the prediction from the encoder and that same prediction fine-tuned for 100 steps.
While using higher order multi-sampling results in a reduction in the LPIPS score for the prediction by the hash estimating encoder and the per-image optimization, there is a small improvement in the score after fine-tuning this prediction.
Given the improvements in the backpropagation provided by this higher order multi-sampling, this is an acceptable tradeoff.
Note that the reduction in the LPIPS score for per-image optimization is expected; the same behaviour was observed in the paper detailing multiresolution hash encoding \cite{muller2022instant}.
These scores also demonstrate the inherent generalizeablity of the HashEncoding architecture.

\begin{table}
  \caption{\textbf{LPIPS scores} of Lagrange multi-sampling with $k = 1$ and $k = 2$ under different optimization strategies averaged across 100 random images from the COCO-val2017 dataset, with $T = 2^{12}$. Note that $k = 1$ is equivalent to bilinear interpolation.}
  \label{tab:percep_loss}
  \centering
  \begin{tabular}{cccc}
    \toprule
    & \multicolumn{3}{c}{Optimization Strategy}                   \\
    \cmidrule(r){2-4}
    $k$ & \thead{Hash Estimating Encoder + \\ Universal Decoder} & \thead{Fine-tuned Encoder Prediction + \\ Universal Decoder}  & \thead{Per-Image Hash Table + \\ Per-Image Decoder} \\
    \midrule
    1 & 0.3622 & 0.2249 & 0.2040 \\
    2 & 0.4107 & 0.2223 & 0.2125 \\
    \bottomrule
  \end{tabular}
\end{table}

\begin{figure}
    \centering
    \begin{subfigure}{0.16\linewidth}
        \includegraphics[width=\linewidth]{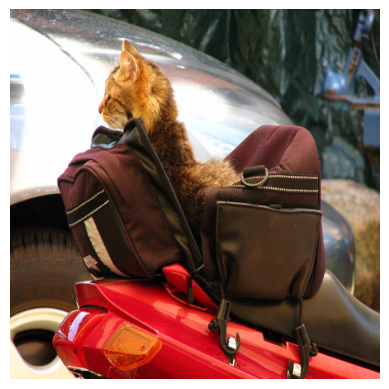}
    \end{subfigure}%
    \begin{subfigure}{0.16\linewidth}
        \includegraphics[width=\linewidth]{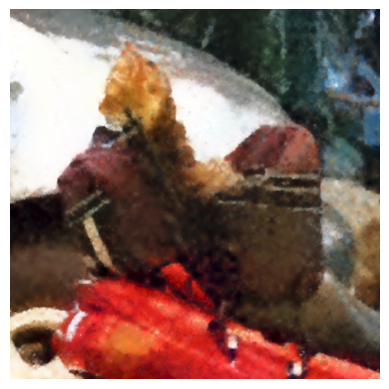}
    \end{subfigure}%
    \begin{subfigure}{0.16\linewidth}
        \includegraphics[width=\linewidth]{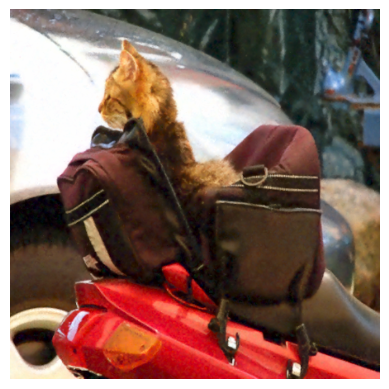}
    \end{subfigure}%
    \begin{subfigure}{0.16\linewidth}
        \includegraphics[width=\linewidth]{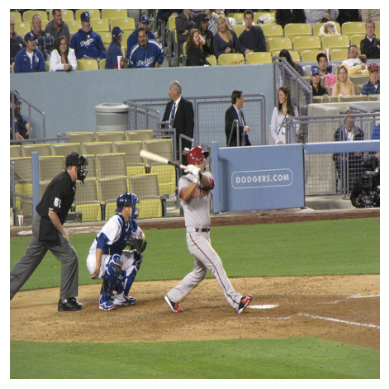}
    \end{subfigure}%
    \begin{subfigure}{0.16\linewidth}
        \includegraphics[width=\linewidth]{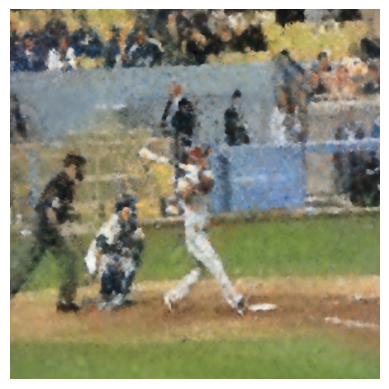}
    \end{subfigure}%
    \begin{subfigure}{0.16\linewidth}
        \includegraphics[width=\linewidth]{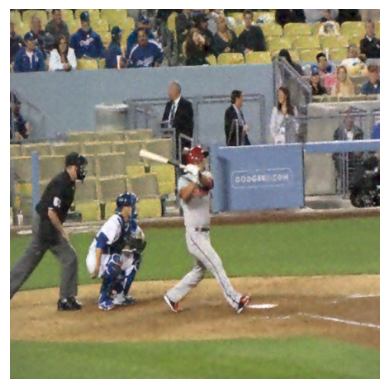}
    \end{subfigure}
    \begin{subfigure}{0.16\linewidth}
        \includegraphics[width=\linewidth]{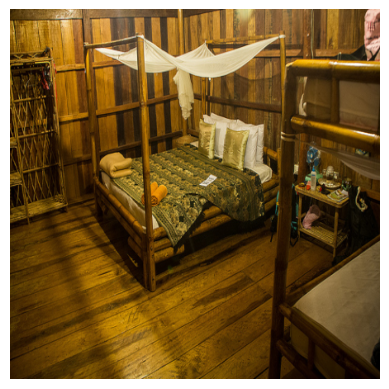}
    \end{subfigure}%
    \begin{subfigure}{0.16\linewidth}
        \includegraphics[width=\linewidth]{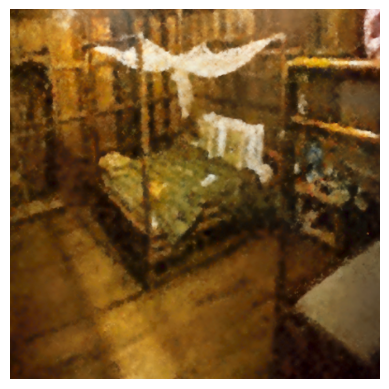}
    \end{subfigure}%
    \begin{subfigure}{0.16\linewidth}
        \includegraphics[width=\linewidth]{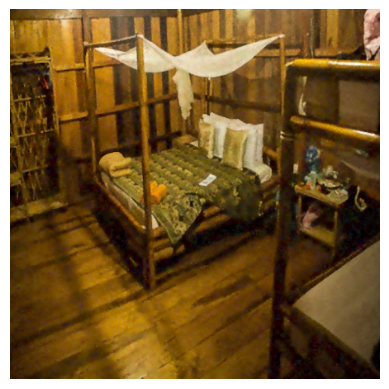}
    \end{subfigure}%
    \begin{subfigure}{0.16\linewidth}
        \includegraphics[width=\linewidth]{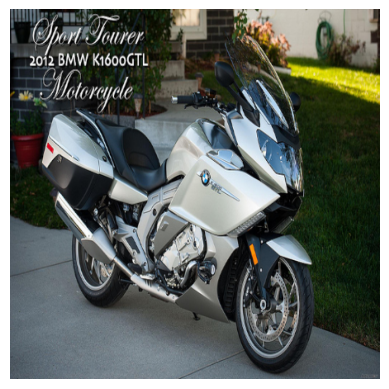}
    \end{subfigure}%
    \begin{subfigure}{0.16\linewidth}
        \includegraphics[width=\linewidth]{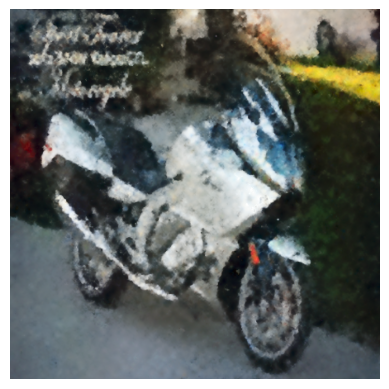}
    \end{subfigure}%
    \begin{subfigure}{0.16\linewidth}
        \includegraphics[width=\linewidth]{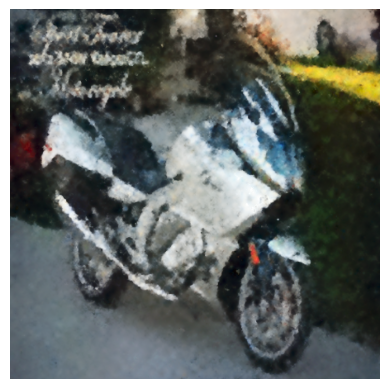}
    \end{subfigure}
    \begin{subfigure}{0.16\linewidth}
        \includegraphics[width=\linewidth]{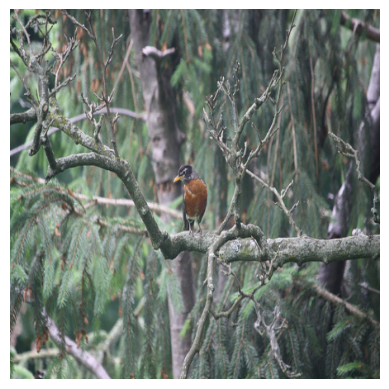}
    \end{subfigure}%
    \begin{subfigure}{0.16\linewidth}
        \includegraphics[width=\linewidth]{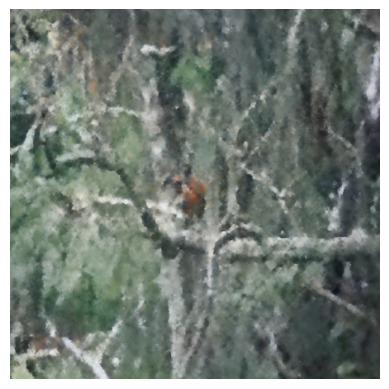}
    \end{subfigure}%
    \begin{subfigure}{0.16\linewidth}
        \includegraphics[width=\linewidth]{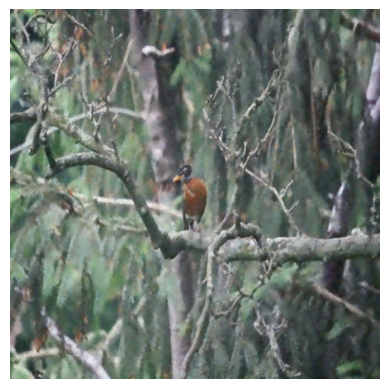}
    \end{subfigure}%
    \begin{subfigure}{0.16\linewidth}
        \includegraphics[width=\linewidth]{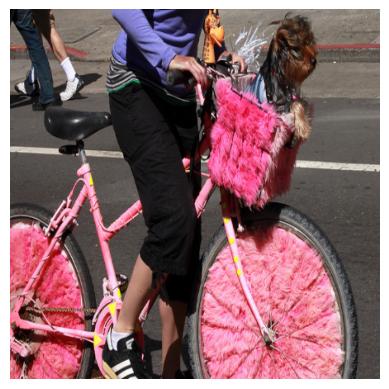}
    \end{subfigure}%
    \begin{subfigure}{0.16\linewidth}
        \includegraphics[width=\linewidth]{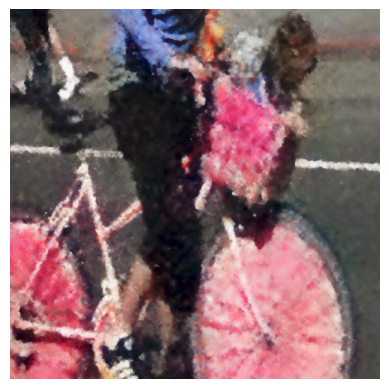}
    \end{subfigure}%
    \begin{subfigure}{0.16\linewidth}
        \includegraphics[width=\linewidth]{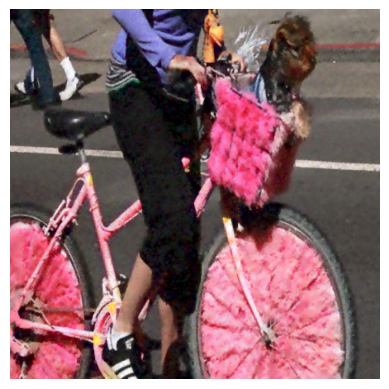}
    \end{subfigure}
    \begin{subfigure}{0.16\linewidth}
        \includegraphics[width=\linewidth]{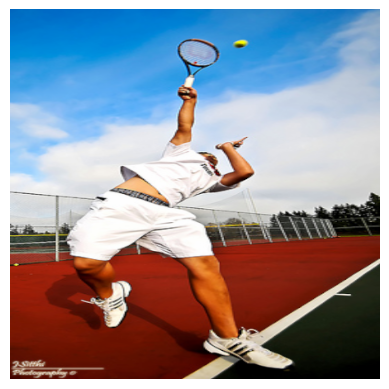}
    \end{subfigure}%
    \begin{subfigure}{0.16\linewidth}
        \includegraphics[width=\linewidth]{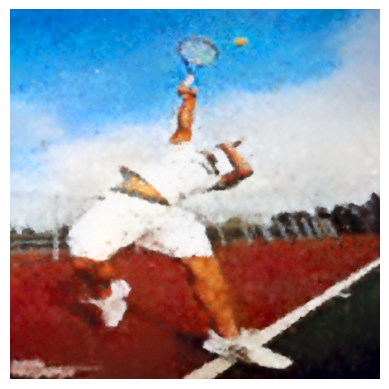}
    \end{subfigure}%
    \begin{subfigure}{0.16\linewidth}
        \includegraphics[width=\linewidth]{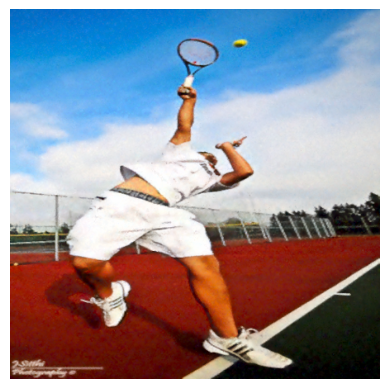}
    \end{subfigure}%
    \begin{subfigure}{0.16\linewidth}
        \includegraphics[width=\linewidth]{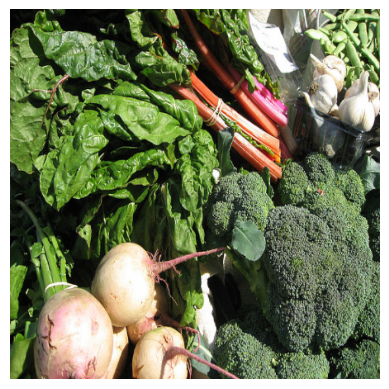}
    \end{subfigure}%
    \begin{subfigure}{0.16\linewidth}
        \includegraphics[width=\linewidth]{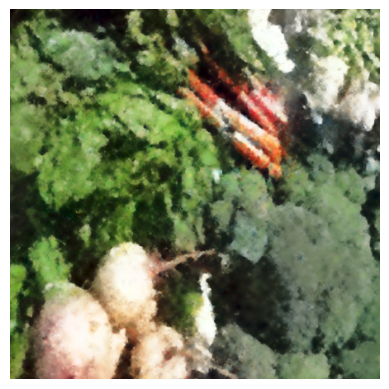}
    \end{subfigure}%
    \begin{subfigure}{0.16\linewidth}
        \includegraphics[width=\linewidth]{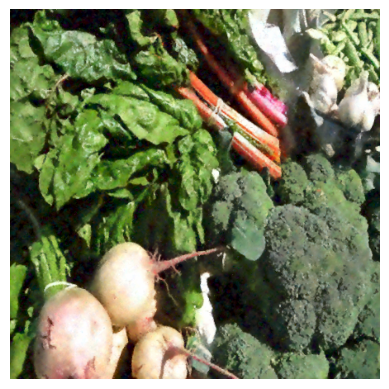}
    \end{subfigure}
    \caption{\textbf{Example reconstruction results.} For each triplet of images, the leftmost is the original image, the middle is the reconstruction predicted from the hash estimating encoder with the universal decoder, and the right is the fine-tuned prediction with the universal decoder.}
    \label{fig:examples}
\end{figure}

\subsection{Translation Invariance of Hash Table}
\begin{figure}[t]
    \centering
    \begin{subfigure}{0.5\linewidth}
        \includegraphics[width=\linewidth]{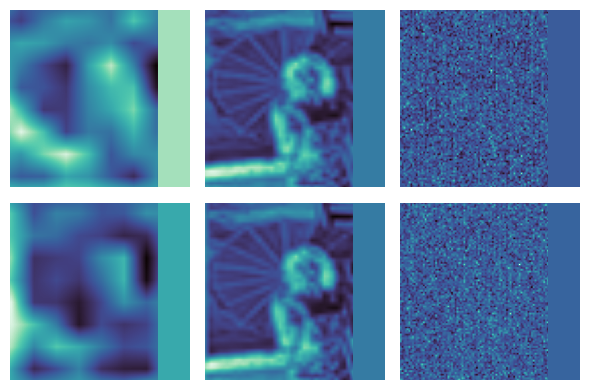}
        \caption{Bilinear}
    \end{subfigure}%
    \begin{subfigure}{0.5\linewidth}
        \includegraphics[width=\linewidth]{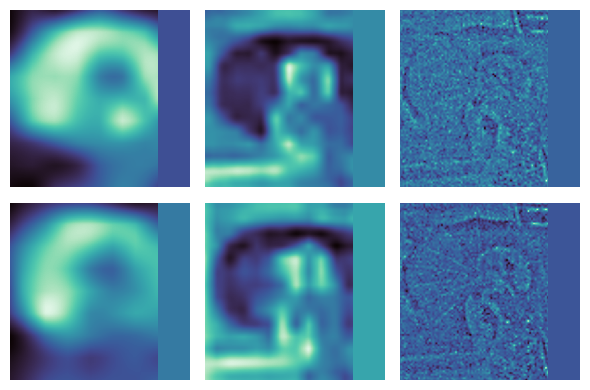}
        \caption{Lagrange Polynomial with k=2}
    \end{subfigure}%
    \caption{\textbf{Heatmap visualizations of translation invariance}. We evaluate the translation invariance with different interpolations/sampling methods, bilinear (left) vs multi-sample (right), and visualize the learned hash features. 
The top row in both (a), (b) visualizes the original untranslated hash table features from $\mathcal{F}(I)$ while the bottom row visualizes the translated and reverted features $\tau^{-1}(\mathcal{F}(\tau(I)))$. To better compare the untranslated encoding with the translated encoding, we zero out the rightmost patch discarded during translation and evaluate the rest. Both models demonstrate a good translation invariance. Note that detailed image features are captured at a finer-hash grid with multi-sampling. }
    \label{fig:invariance}
\end{figure}
Convolutional neural network leverages its translation invariance to achieve better generalization in computer vision tasks. We empirically reveal the same property of multi-resolution hash table, shedding light on further exploitation of non-parametric data structures. Given a universal translation function $\tau$, the hash table $\mathcal{F}$ is translation invariant if $\mathcal{F}(I) = \tau^{-1}(\mathcal{F}(\tau(I)))$. Namely, we can reconstruct $I$ by shifting back the encoding from the hash table defined by the shifted image $\tau(I)$. Figure \ref{fig:invariance} demonstrates the translation invariance under different interpolation settings by visualizing intermediate features $[4,12,16]$ out of the total 24 ($L=12, F=2$) as heatmaps. In both \ref{fig:invariance}(a) and \ref{fig:invariance}(b), important features are retained in earlier layers of the hash table, from coarse to detailed, and we can observe the translation invariance by comparing untranslated original hash table at the top row and the translated at the bottom, where little impact from translation applies. Note that the latter features collapse, despite the better performance of multisampling in \ref{fig:invariance}(b) which enables them to collapse slower. Because of the space-folding property of the hash table, it is hard to find consensus on the alignment of the finest details. We would need to depend on fine-tuning the decoder to help us restore the finer details.


\subsection{Optical Flow using gradients of coordinates}
One natural task for coordinate-based image synthesis is to estimate spatial image transformation by inverting the synthesis pipeline.   Because our pixel decoder is extremely simple and learned hash features are translation invariant, we conjunct that a simple SGD backpropagation is sufficient to achieve optical flow computation without using any encoder.

Given a coordinate $(x, y)$ in the original image $I_1$, we seak the $(\Delta x, \Delta y)$ such that $(x + \Delta x, y + \Delta y)$ is its corresponding pixel in the transformed image $I_2$ with $I_1(x, y) = I_2(x + \Delta x, y+\Delta y)$. We first encode both images in their hash encoding.  For each pixel coordinate $(x,  y)$ in $I_1$, we optimize a $(\Delta x, \Delta y)$ by back-propagating the $L_2$ loss between the decoded pixel value $I'_1(x, y)$ and $I'_2(x+\Delta x, y + \Delta y)$, where $I'_1, I'_2$ denotes the decoded image from the hash tables we encoded.    

We carried out a simple experiment on Coco dataset. We take 100 images from the dataset to form the test set. For each image, we randomly translate the whole image  $x$ and $y$ direction within radius of 50 pixels.   We minimize the pixel loss for 256 randomly sampled points, discarding those within 50 pixels from the image boundaries.
For the patch-based matching, we form a $3\times3$ patch of the pixels by adding or subtracting 1 pixel in $x$ and $y$ coordinates.  
We constraint that all the pixel coordinates share the same $(\Delta x, \Delta y)$, within the whole image or patch. We compare the EPE (end-point error) of the optical flow for all the methods using both bilinear interpolation and high-order Lagrange Polynomial interpolation. The results shown in Table ~\ref{tab:opticalflow} demonstrates that higher-order Lagrange Polynomial interpolation provides better results in all three methods. This is because larger spatial translations require gradient propagation over a larger neighborhood across multiple grid cell boundaries.   Higher-order Lagrange Polynomial produces a smoother gradient across these grid boundaries, providing more stability in back-propagation.

\begin{table}
  \caption{\textbf{EPE} of the optical flow on image translations. Lagrange multi-sampling with $k = 1$ and $k = 2$ under different methods averaged across 100 random images from the COCO-eval dataset. Note that $k = 1$ is equivalent to bilinear interpolation.}
  \label{tab:opticalflow}
  \centering
  \begin{tabular}{cccc}
    \toprule
    &\multicolumn{3}{c}{Optical flow Method}                   \\
    \cmidrule(r){2-4}
    $k$ & \thead{Pixel-wise} & \thead{Patch-wise}  & \thead{Image-wise} \\
    \midrule
    1 & 36.4531 & 34.0380 & 4.2565 \\
    2 & 34.5785 & 33.0150 & 2.9412 \\
    \bottomrule
  \end{tabular}
\end{table}



\section{Conclusion}\vspace{-5pt}

 We developed a universal image encoding method based on a non-parametric multiscale coordinate hash function to achieve a per-pixel decoder without convolutions, inspired by the work of \cite{muller2022instant}.  In contrast to \cite{muller2022instant}, which is per-image based, our method produces a universal decoder/hash representation to a potentially infinite number of images.   We leverage the space-folding behavior of hashing functions to find recurrent image features across image scales by building on a SegFormer architecture.  We introduce multi-sampling, which allows effective backpropagation directly to the coordinate space.   We show that our method, called HashEncoding, can be exploited for geometric tasks such as optical flow. 

\FloatBarrier
\bibliography{egbib}

\appendix


\newpage
\def\B#1{\textcolor{blue}{#1}}

\def\R#1{\textcolor{red}{#1}}

\begin{wrapfigure}{L}{0.4\linewidth}
    \vspace{2em}
    \includegraphics[width=\linewidth]{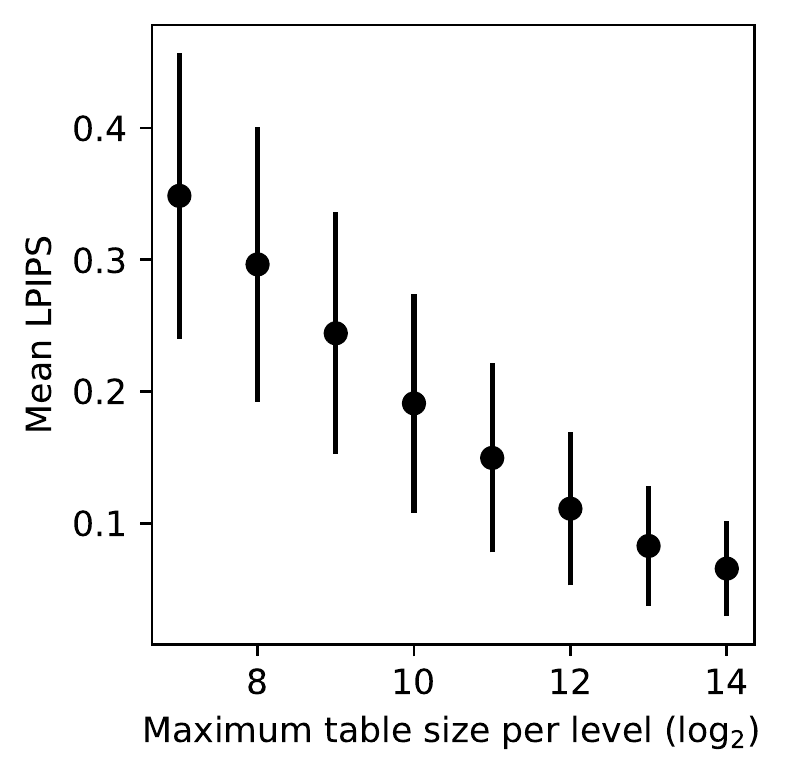}
    \caption{\textbf{Mean LPIPS score for different maximum hash table sizes} across 100 images from the COCO-val2017 dataset, optimized per-image with standard deviation shown.}
    \label{fig:hsize_lpips}
\end{wrapfigure}

\begin{figure}
    \captionsetup{justification=centering}
    \begin{subfigure}{0.5\linewidth}
        \includegraphics[width=\linewidth]{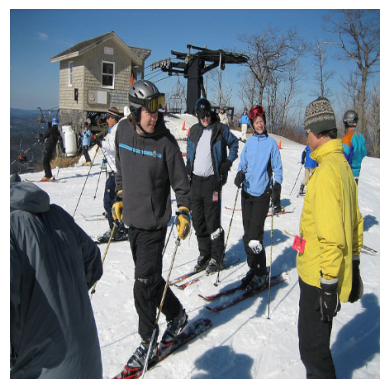}
        \caption{Original}
    \end{subfigure}%
    \begin{subfigure}{0.5\linewidth}
        \includegraphics[width=\linewidth]{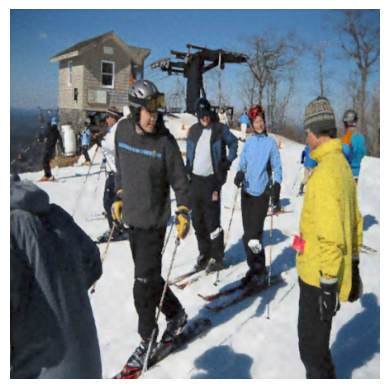}
        \caption{Reconstruction}
    \end{subfigure}
    \begin{subfigure}{0.5\linewidth}
        \includegraphics[width=\linewidth]{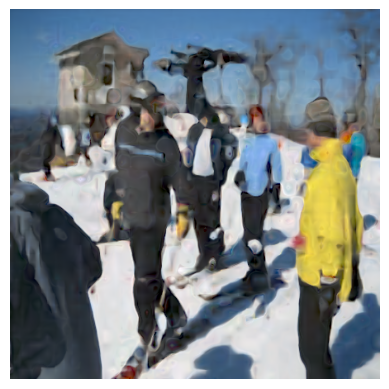}
        \caption{Only layers without hash collisions}
    \end{subfigure}%
    \begin{subfigure}{0.5\linewidth}
        \includegraphics[width=\linewidth]{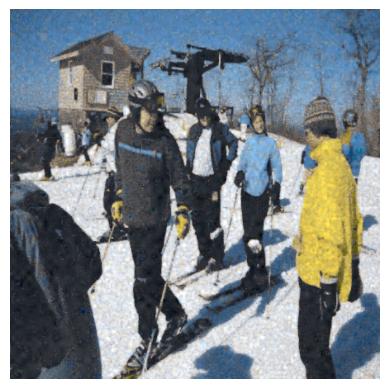}
        \caption{Only layers with hash collisions}
    \end{subfigure}
    \captionsetup{justification=justified}
    \caption{\textbf{Effect of layers with and with hash collision} on per-image reconstruction with $T = 2^{12}$. Here, layers without hash collisions are the coarsest levels of the hash table that do not repeat entries while layers with hash collisions are the finer layers that exhibit a repeating hash structure. Compared to the original image, b, c and d represent and LPIPS of 0.1471, 0.3200 and 0.2573, respectively.}
    \label{fig:drop_layers}
\end{figure}

\section{Structure of the Hash Table}\vspace{-5pt}

As shown in Figure \ref{fig:hsize_lpips}, a larger hash table allows for a better reconstruction of the original image.
However, a larger hash table comes at the expense of a larger embedding space.
Given our use-case of optical flow, we choose a hash table size to maximize image quality while being smaller than the original image.
With a maximum table size of $2^{12}$ per level, this results in an embedding space that is approximately half the size of the (compressed) original image on disk. 

As shown in Figure \ref{fig:hash_table}, the coarsest layers of the hash table do not have repeating entries while the finer layers show a regular, hashed pattern.
For a maximum table size of $2^{12}$ per level and 12 levels, the first 7 levels (up to a resolution of $46 \times 46$) show no repetition.
One might question if both these types of layers are necessary or if one encodes most of the information.
Figure \ref{fig:drop_layers} demonstrates that both types of layers are necessary to achieve suitable reconstruction.
With only dense layers, the resulting reconstruction appears blurry, while only hashed layers produces far too much noise and loses colour detail.

More detailed analysis needs to be conducted to determine the optimal structure of the hash table for different tasks.

\section{Ablation Study}\vspace{-5pt}
\subsection{Encoders}\label{ablation}
\begin{table}
  \caption{\textbf{LPIPS scores}$(\downarrow)$ of using different encoders under different optimization strategies averaged across 100 random images from the COCO-val2017 with $T = 2^{12}$, following \ref{recon_quality} and extending \ref{tab:percep_loss}. For each optimization strategy, we highlight the best with blue and the second best with red.}
  \label{tab:extend_percep_loss}
  \centering
  \begin{tabular}{ccccc}
    \toprule
    && \multicolumn{3}{c}{Optimization Strategy}                   \\
    \cmidrule(r){3-5}
    \thead{Encoder + \\ n. params } & $k$ & \thead{Hash Estimating Encoder + \\ Universal Decoder} & \thead{Fine-tuned Encoder Prediction + \\ Universal Decoder}  & \thead{Per-Image Hash Table + \\ Per-Image Decoder} \\
    \midrule
    MiT-B0\cite{xie2021segformer}  &1 & \B{0.3622 $\pm$ 0.11} & 0.2249 $\pm$ 0.08 & 0.2040 $\pm$ 0.05\\
    3.3M        &2 & \R{0.4107 $\pm$ 0.08} & \R{0.2223 $\pm$ 0.06} & 0.2125 $\pm$ 0.06\\
    \midrule
    U-Net\cite{ronneberger2015u}  &1 & 0.4622 $\pm$ 0.09 & 0.2270 $\pm$ 0.06 & \R{0.1961 $\pm$ 0.05}\\ 
    487M        &2 & 0.4706 $\pm$ 0.09 & \B{0.2203 $\pm$ 0.06} & 0.2347 $\pm$ 0.07\\
    \midrule
    ResNet\cite{he2016deep}  &1 & 0.7156 $\pm$ 0.12 & 0.4420 $\pm$ 0.10 & \B{0.1959 $\pm$ 0.05}\\
    36.2M        &2 & 0.6962 $\pm$ 0.12 & 0.3268 $\pm$ 0.08 & 0.2134 $\pm$ 0.05\\ 
    \bottomrule
  \end{tabular}
\end{table}
We evaluate the effectiveness of different encoders w.r.t model size and architecture, under different optimization strategies. We follow and extend Table \ref{tab:percep_loss} to compare Mix Transformer encoder with FCN based encoders. While increasing model size and adopting intermediate connections improve the initial hash estimations for FCN encoders, a Mix Transformer can outperform FCN encoders by a large margin with much fewer parameters. Both universal fine-tuning and per-image fine-tuning can further improve the performance of different encoders. For fine-tuning encoder prediction, Mit-B0 and U-Net can be improved to the same level with U-Net ($k=2$) slightly better and both outperforms ResNet. For per-image hash table and decoder, the impact of encoder choice will be minimized: All encoders in our settings can be tuned to the same extent, with ResNet ($k=1$) slightly superior. The adaptation to multi-sampling can improve ResNet under universal decoder settings, but the effect is too subtle for other two encoders, or even harmful to the reconstruction quality of the Mix Transformer encoder model, since the larger spatial support of multi-sampling will compound small errors produced by the Mix Transformer.

\subsection{Quantitative Measurement of Translation Invariance}
\begin{table}
  \caption{\textbf{Evaluating translation invariance} based on the divergence of hash table encoding$\mathcal{F}(I)$ from its shifted and recovered version $\tau^{-1}(\mathcal{F}(\tau(I)))$ under different optimization strategies and encoders, measured with LPIPS scores. We calculate the average scores across 100 random images from the COCO-val2017 dataset, with $T = 2^{12}$ and fixed random seed for reproducibility.}
  \label{tab:quant_invar}
  \centering
  \begin{tabular}{ccc}
    \toprule
    Encoder & $k=1$ & $k=2$ \\
    \midrule
    MiT-B0 & 0.0337 & 0.0891 \\
    U-Net & 0.1183 & 0.1207 \\
    ResNet & 0.0597 & 0.0579 \\
    \bottomrule
  \end{tabular}
\end{table}
In addition to visualizations in Figure \ref{fig:invariance}, we can also quantitatively measure and compare the translation invariance of various models pre-trained under different optimization strategies using LPIPS scores between $\mathcal{F}(I)$ and $\tau^{-1}(\mathcal{F}(\tau(I)))$. We report the average scores across 100 random images from validation set for each experiment setting in Table \ref{tab:quant_invar}. At each iteration, we perform a random translation along x-axis of image by $r$ pixels where $r\in[-80, 80]$ and is a multiple of 10. While all the models can achieve quite low LPIPS socres and thus demonstrating translation invariance, we generally observe that Mix Transformer encoder achieves better invariance than FCN encoders. The gain for applying multi-sampling is subtle under this context, and will compound errors when adopting Mix Transformer, which is consistent with the previous observations in \ref{ablation}.

\section{Additional Visualizations}\vspace{-5pt}
\begin{figure}
    \centering
    \begin{subfigure}{0.5\linewidth}
        \includegraphics[width=\linewidth]{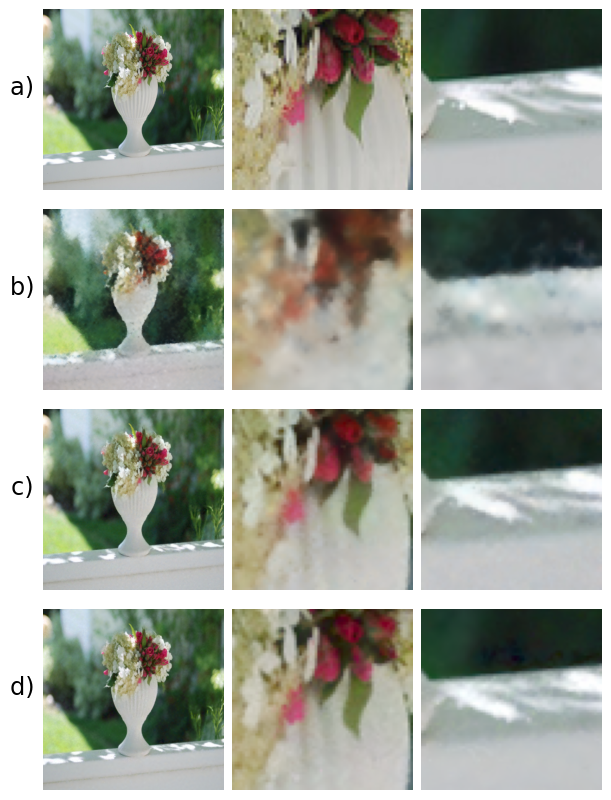}
    \end{subfigure}%
    \begin{subfigure}{0.5\linewidth}
        \includegraphics[width=\linewidth]{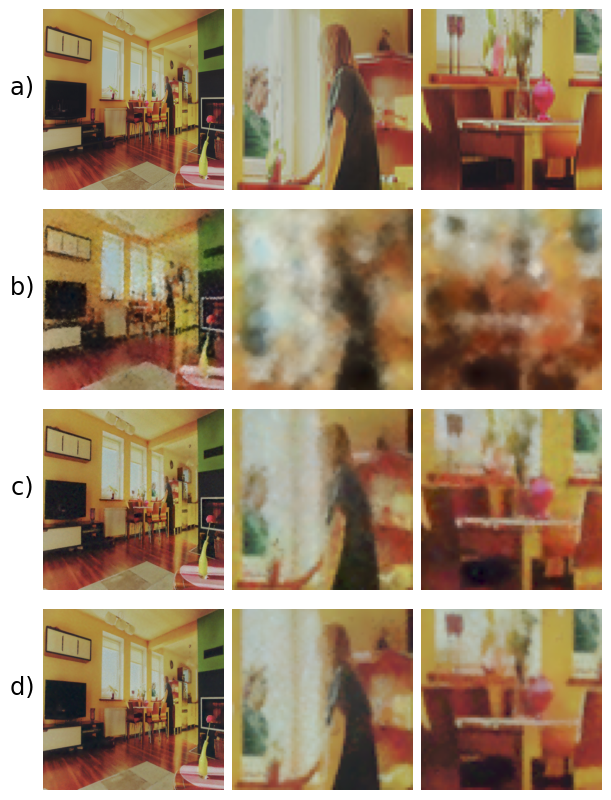}
    \end{subfigure}
    \begin{subfigure}{0.5\linewidth}
        \includegraphics[width=\linewidth]{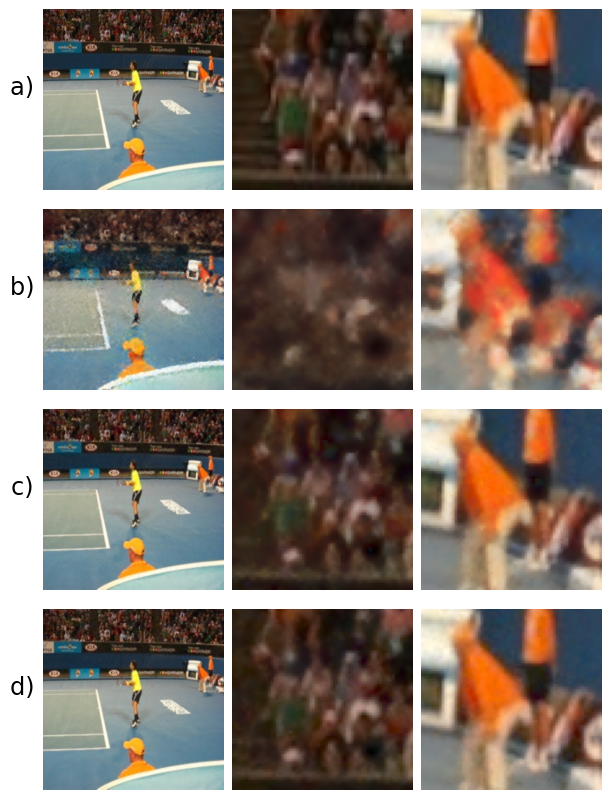}
    \end{subfigure}%
    \begin{subfigure}{0.5\linewidth}
        \includegraphics[width=\linewidth]{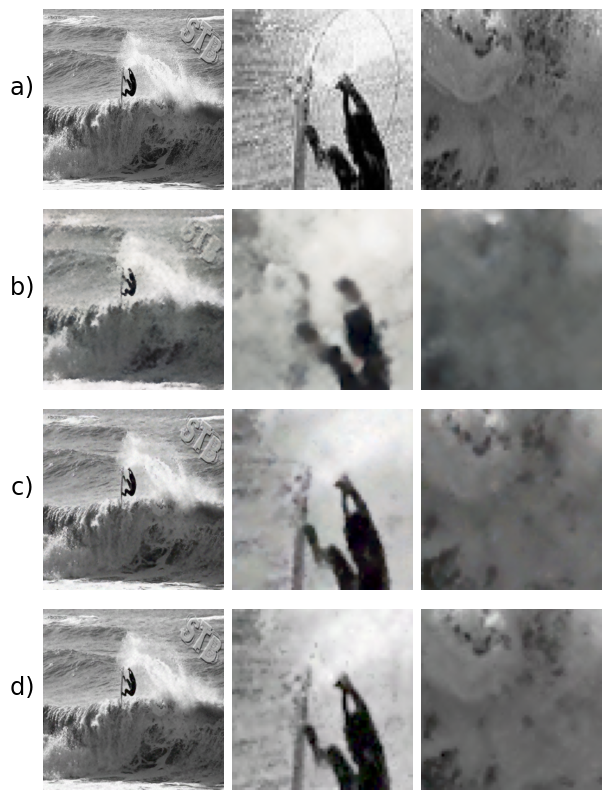}
    \end{subfigure}
    \caption{\textbf{Additional examples with zoom-in.} For each image, we show \textbf{a)} the original image, \textbf{b)} the reconstruction using the hash estimating encoder with the universal decoder, \textbf{c)} the fine-tuned reconstruction with the universal decoder, and \textbf{d)} the per-image reconstruction.}
    \label{fig:examples1}
\end{figure}

\begin{figure}
    \centering
    \begin{subfigure}{0.5\linewidth}
        \includegraphics[width=\linewidth]{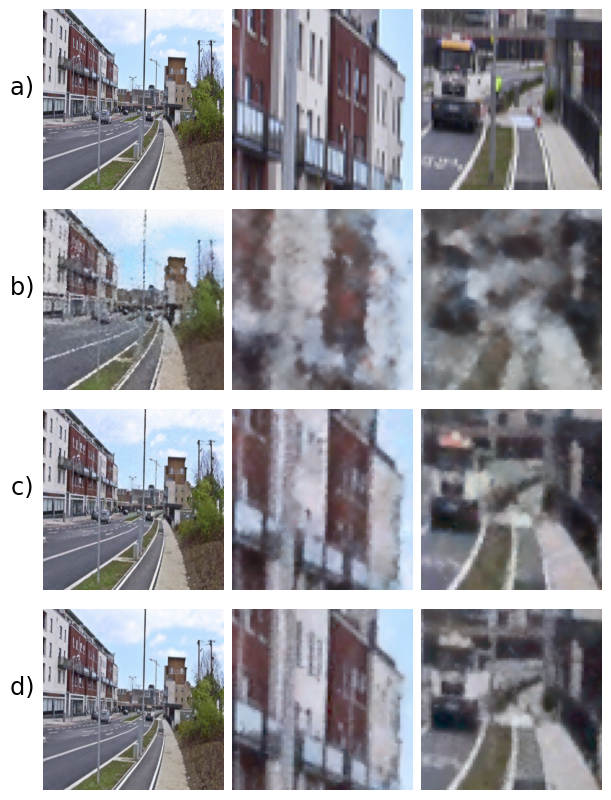}
    \end{subfigure}%
    \begin{subfigure}{0.5\linewidth}
        \includegraphics[width=\linewidth]{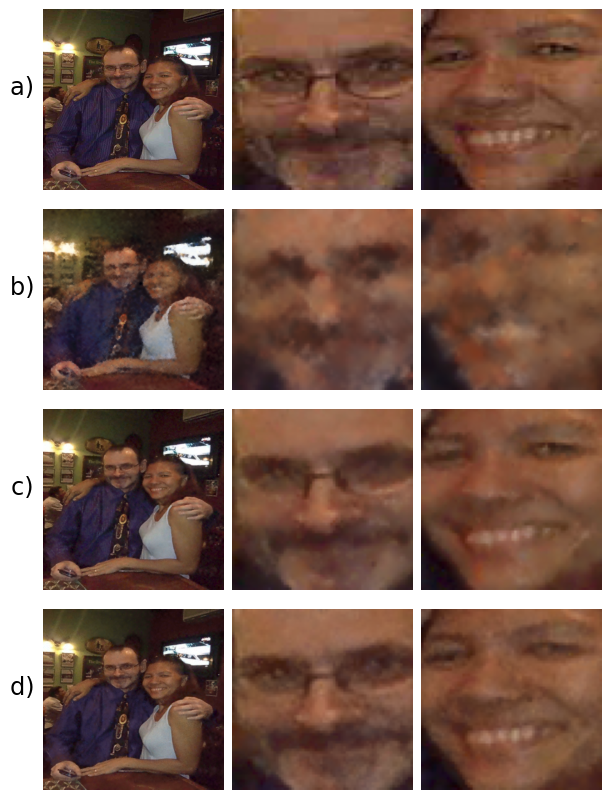}
    \end{subfigure}
    \begin{subfigure}{0.5\linewidth}
        \includegraphics[width=\linewidth]{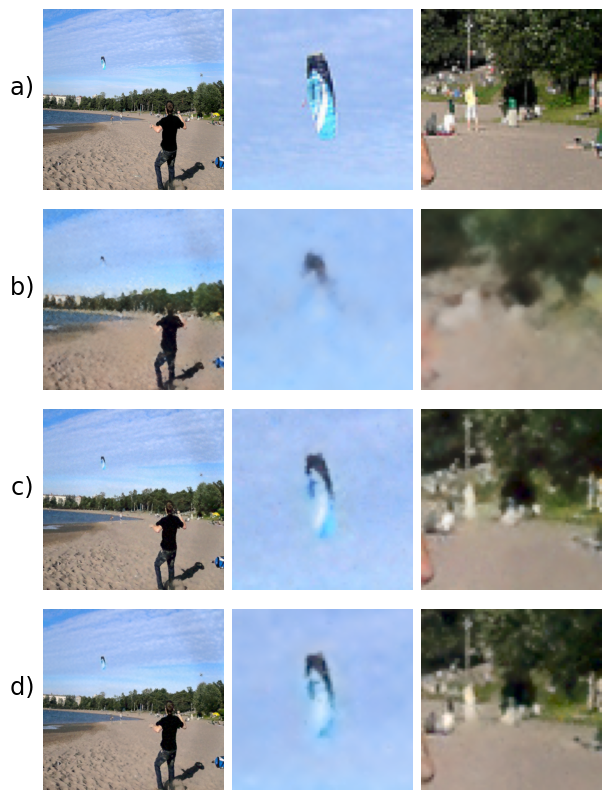}
    \end{subfigure}%
    \begin{subfigure}{0.5\linewidth}
        \includegraphics[width=\linewidth]{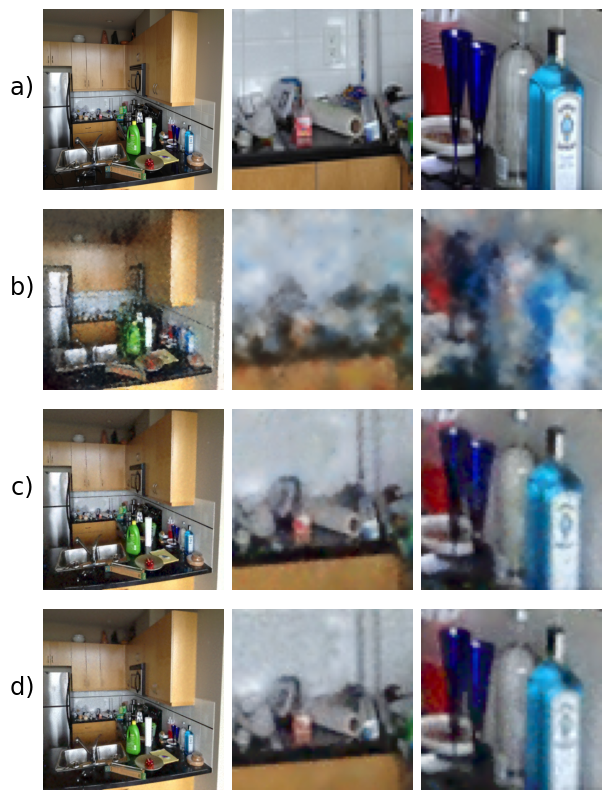}
    \end{subfigure}
    \caption{\textbf{Additional examples with zoom-in (cont).} For each image, we show \textbf{a)} the original image, \textbf{b)} the reconstruction using the hash estimating encoder with the universal decoder, \textbf{c)} the fine-tuned reconstruction with the universal decoder, and \textbf{d)} the per-image reconstruction.}
    \label{fig:examples2}
\end{figure}

\begin{figure}
    \centering
    \includegraphics[width=\linewidth]{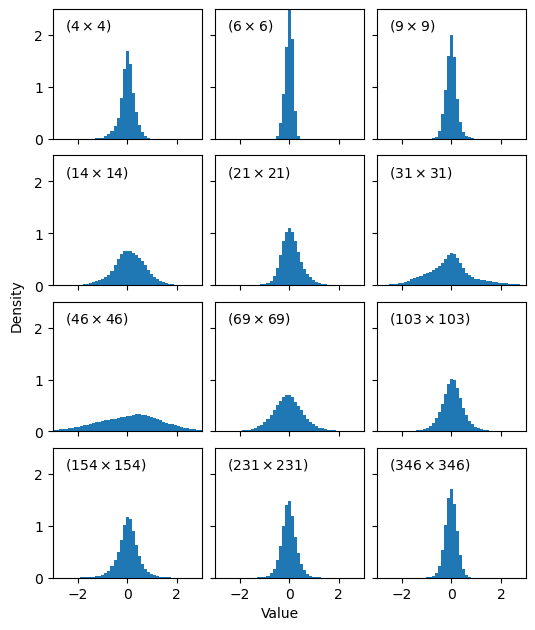}
    \caption{\textbf{Density histogram of entries in each level of the hash table} across 100 images from the COCO-val2017 dataset. The grid resolution associated with each histogram is shown. The entries in each layer appear to be zero-centered and symmetric.}
    \label{fig:dens_hist}
\end{figure}

\begin{figure}
    \centering
    \begin{subfigure}{\linewidth}
        \includegraphics[width=\linewidth]{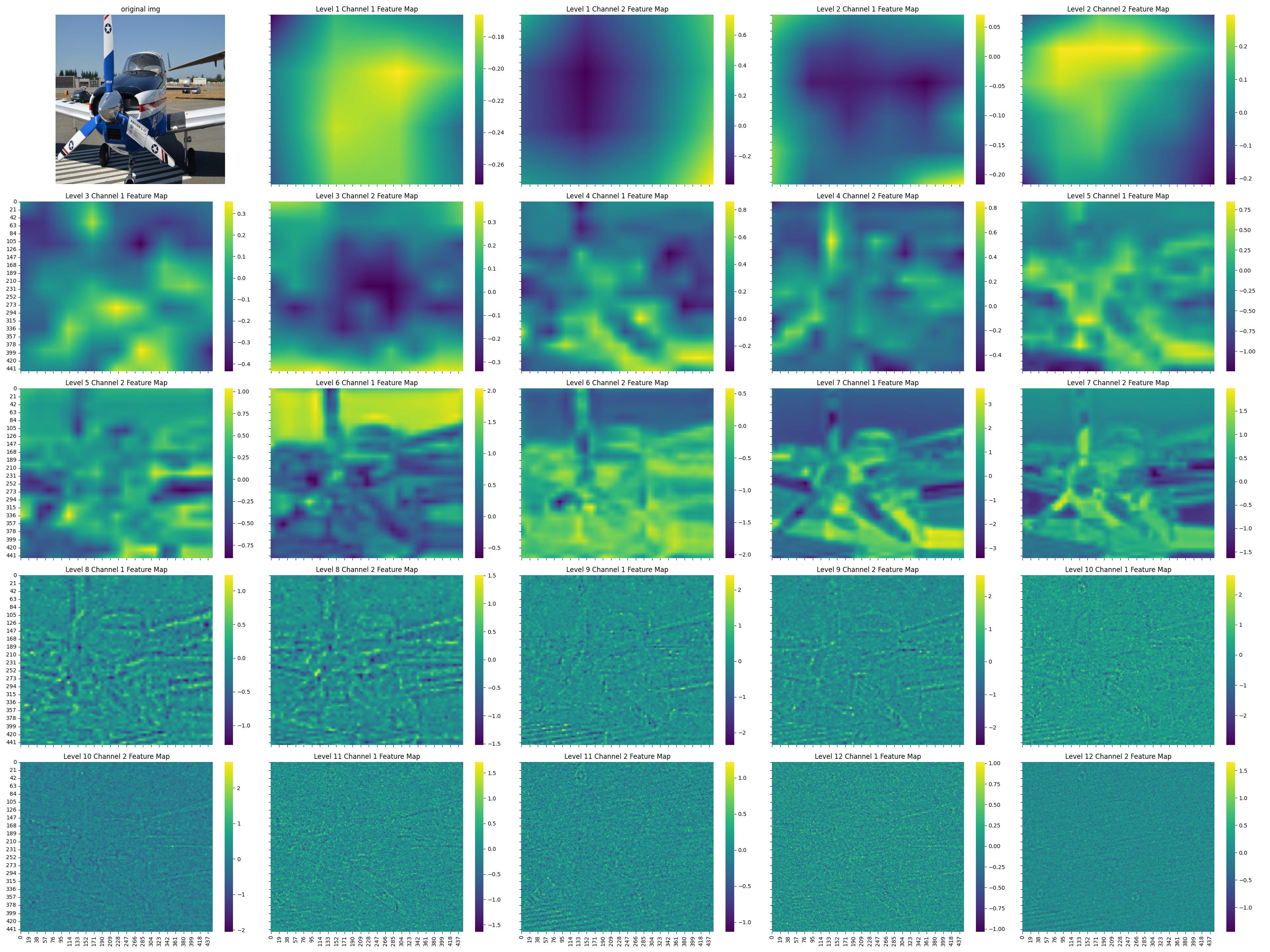}
    \end{subfigure}
    \begin{subfigure}{\linewidth}
        \includegraphics[width=\linewidth]{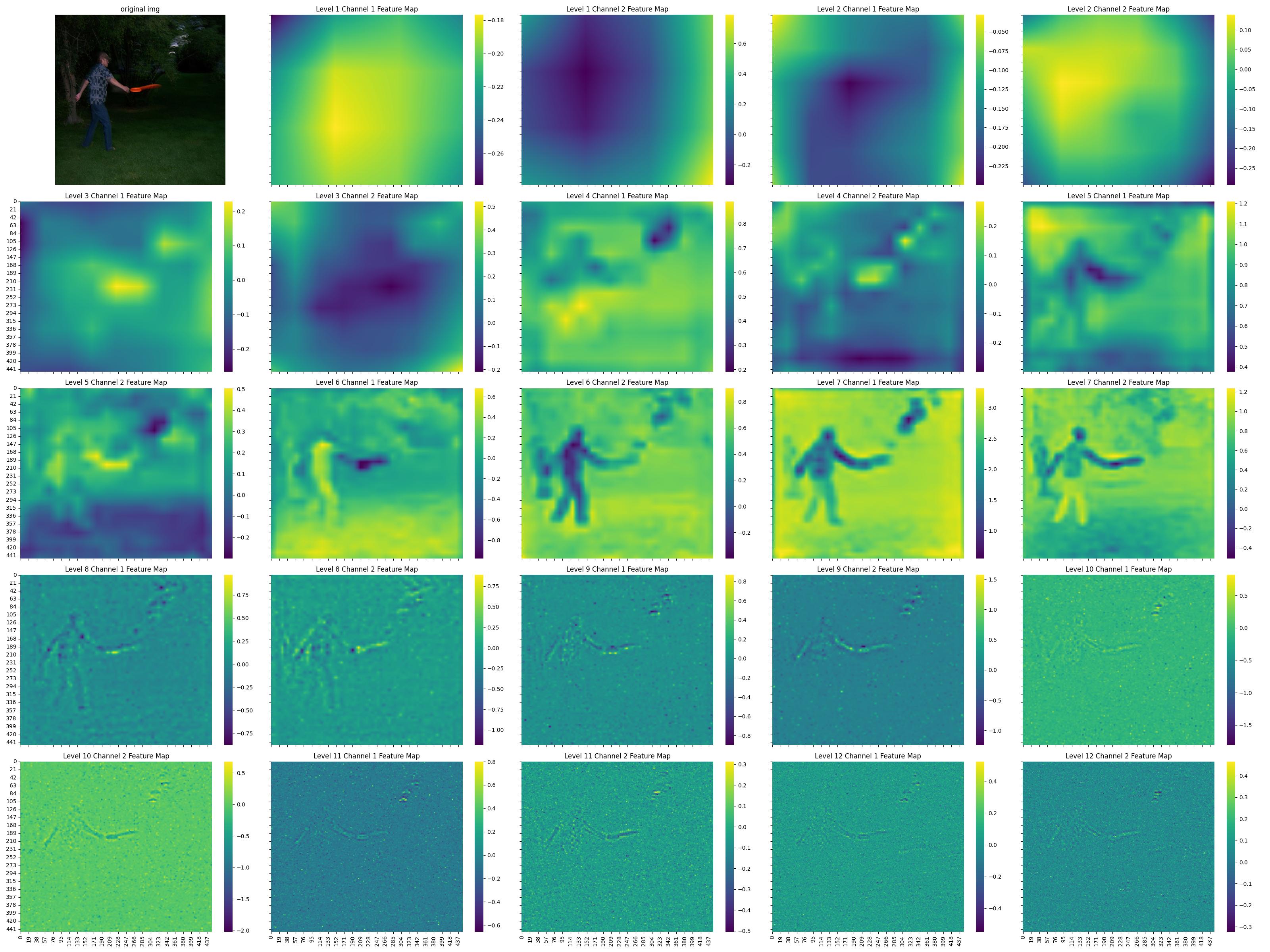}
    \end{subfigure}
    \caption{\textbf{Feature maps generated from different levels of hashmaps.} Feature maps from low level of hashtable to high level are shown from top left to bottom right. }
    \label{fig:featmap1}
\end{figure}

\begin{figure}
    \centering
    \begin{subfigure}{\linewidth}
        \includegraphics[width=\linewidth]{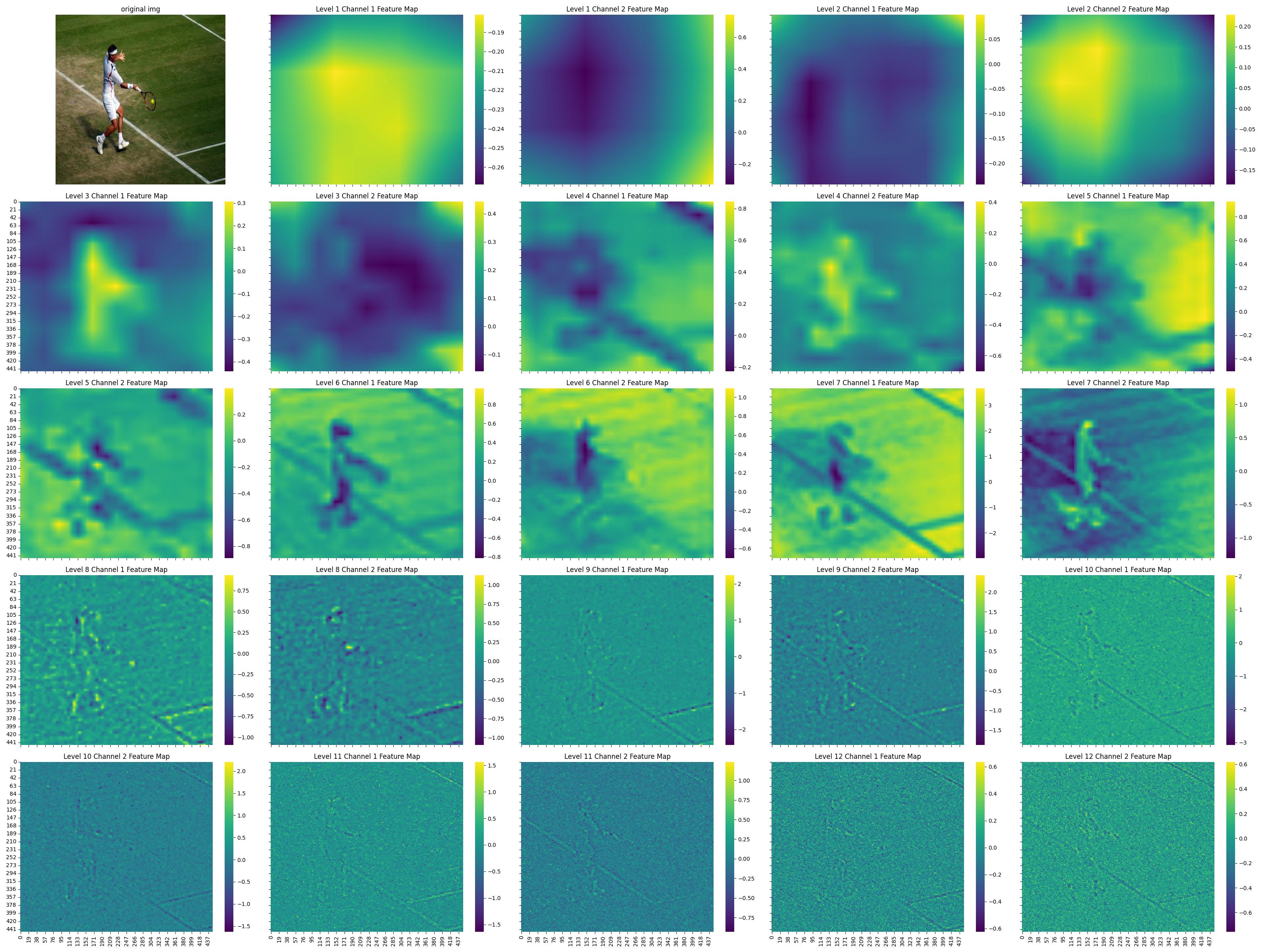}
    \end{subfigure}
    \begin{subfigure}{\linewidth}
        \includegraphics[width=\linewidth]{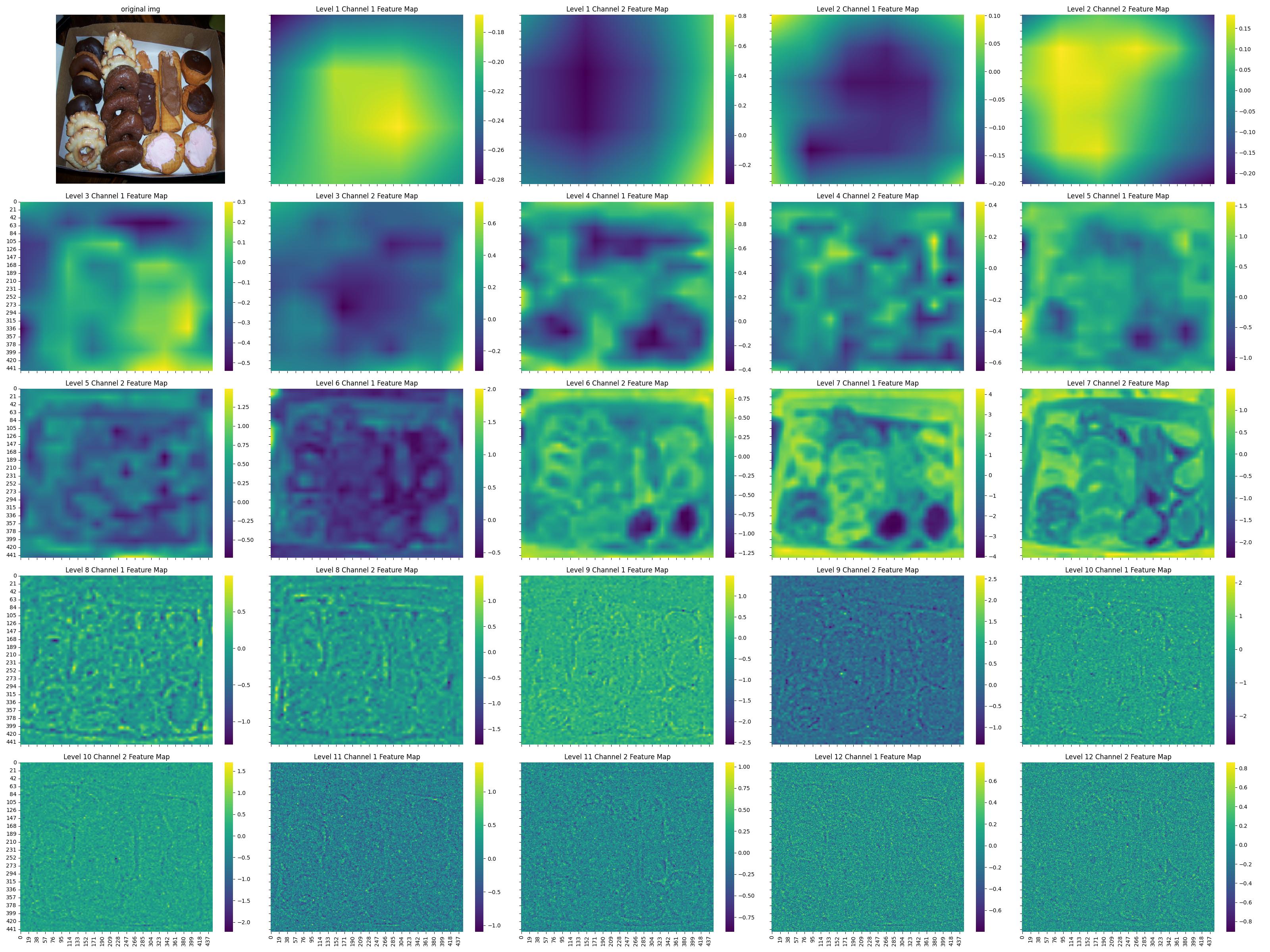}
    \end{subfigure}
    \caption{\textbf{More feature maps.}}
    \label{fig:featmap2}
\end{figure}

\end{document}